\documentclass[letterpaper,twocolumn,10pt]{article}
\usepackage{usenix-2020-09}

\usepackage[normalem]{ulem}

\usepackage{amsmath}

\usepackage{xspace}
\usepackage[inline]{enumitem}
\usepackage{soul}
\usepackage{xcolor}
\usepackage{comment}
\usepackage{textcomp}
\usepackage{tcolorbox}
\usepackage{authblk}
\usepackage[hang,flushmargin]{footmisc} 

\makeatletter
\renewcommand\AB@affilsepx{\qquad\protect\Affilfont}
\makeatother

\usepackage{url}

\usepackage{float}
\usepackage{graphicx}

\usepackage{tabularx, booktabs}

\usepackage{xspace}
\usepackage{listings}
\usepackage{placeins}
\usepackage{makecell}

\usepackage{tikz}
\usepackage{pgfplots}
\usepackage{pgfplotstable}
\pgfplotsset{compat=1.18}
\usepgfplotslibrary{colorbrewer}
\usepgfplotslibrary{dateplot}

\usetikzlibrary{positioning}
\usepackage{multirow}
\usepackage{setspace}
\usepackage{subcaption}
\usepackage{tablefootnote}
\usepackage{pifont}
\usepackage{balance}
\usepackage{soul}

\usepackage[capitalise,nameinlink]{cleveref}
\crefformat{section}{#2\S#1#3}
\Crefformat{section}{#2\S#1#3}
\crefformat{subsection}{#2\S#1#3}
\Crefformat{subsection}{#2\S#1#3}
\crefname{line}{Line}{Lines}
\Crefname{line}{Line}{Lines}
\crefformat{line}{#2Line~#1#3}
\Crefformat{line}{#2Line~#1#3}
\crefname{listing}{Listing}{Listings}
\Crefname{listing}{Listing}{Listings}
\crefname{lstlisting}{Listing}{Listings}
\Crefname{lstlisting}{Listing}{Listings}
\crefformat{listing}{#2Listing~#1#3}
\Crefformat{listing}{#2Listing~#1#3}
\crefname{figure}{Figure}{Figures}
\Crefname{figure}{Figure}{Figures}
\crefformat{figure}{#2Fig.~#1#3}
\Crefformat{figure}{#2Fig.~#1#3}

\crefrangeformat{section}{#3\S#1#4--#5#2#6}
\Crefrangeformat{section}{#3\S#1#4--#5#2#6}
\crefrangeformat{subsection}{#3\S#1#4--#5#2#6}
\Crefrangeformat{subsection}{#3\S#1#4--#5#2#6}

\crefname{table}{Table}{Tables}
\Crefname{table}{Table}{Tables}

\usepackage{listings}
\usepackage{xcolor}
\usepackage{soul}

\definecolor{yelloworange}{RGB}{255, 200, 0}
\sethlcolor{yelloworange}

\definecolor{darkgreen}{RGB}{0, 128, 0}
\definecolor{commentgreen}{rgb}{0, 0.5, 0}

\lstdefinelanguage
[x64]{Assembler}     %
[x86masm]{Assembler} %
{morekeywords={CDQE,CQO,CMPSQ,CMPXCHG16B,JRCXZ,LODSQ,MOVSXD, %
		POPFQ,PUSHFQ,SCASQ,STOSQ,IRETQ,RDTSCP,SWAPGS, %
		rax,rdx,rcx,rbx,rsi,rdi,rsp,rbp, %
		r8,r8d,r8w,r8b,r9,r9d,r9w,r9b}} %

\lstdefinestyle{customc}
{
	belowcaptionskip=-0.5\baselineskip,
	breaklines=true,
	captionpos=b,                    %
	language=C,
	showstringspaces=false,
	basicstyle=\fontsize{8}{7}\selectfont\bfseries\ttfamily,
	keywordstyle=\color{black},
	commentstyle=\itshape\color{gray!70!black},
	identifierstyle=\color{black},
	stringstyle=\color{red!70!black},
	emph={static,volatile,double,float,signed,unsigned,int,void,size_t,char, key_t, value_t,STORE, FLUSH,FENCE,uint64_t,struct},
	emphstyle={\color{olive}},
	numbersep=8pt,
}

\lstdefinestyle{customnew}
{
    backgroundcolor=\color{lightgray!10}, %
    commentstyle=\color{darkgreen}\textit, %
    keywordstyle=[1]\color{blue}\bfseries, %
    keywordstyle=[2]\color{purple}, %
    numberstyle=\tiny\color{gray}, %
    stringstyle=\color{orange}, %
    basicstyle=\ttfamily\footnotesize, %
    breakatwhitespace=false, %
    breaklines=true, %
    captionpos=b, %
    keepspaces=true, %
    numbers=left, %
    numbersep=5pt, %
    showspaces=false, %
    showstringspaces=false, %
    showtabs=false, %
    tabsize=2, %
    frame=single, %
    rulecolor=\color{black}, %
    aboveskip=1.5em, %
    belowskip=1.5em, %
    frameround=tttt, %
    morekeywords=[2]{np,zeros,mean,std,ndarray}, %
}

\lstdefinestyle{customblackwhite}
{
    backgroundcolor=\color{white}, %
    commentstyle=\color{darkgreen}\textit, %
    keywordstyle=[1]\color{blue}\bfseries, %
    keywordstyle=[2]\color{purple}, %
    numberstyle=\tiny\color{gray}, %
    stringstyle=\color{orange}, %
    basicstyle=\ttfamily\footnotesize, %
    breakatwhitespace=false, %
    breaklines=true, %
    captionpos=b, %
    keepspaces=true, %
    numbers=left, %
    numbersep=5pt, %
    showspaces=false, %
    showstringspaces=false, %
    showtabs=false, %
    tabsize=2, %
    frame=single, %
    rulecolor=\color{black}, %
    aboveskip=1.5em, %
    belowskip=1.5em, %
    frameround=tttt, %
    morekeywords=[2]{np,zeros,mean,std,ndarray}, %
}

\usepackage{algorithm}
\usepackage{algpseudocode}

\newif\ifdraft
\draftfalse

\newcommand{\sys}[1]{\textsc{Argos}}
\newcommand{\msft}[1]{{Microsoft}}

\newcommand{\stt}[1]{\texttt{\small #1}\xspace}

\renewcommand{\paragraph}[1]{\textbf{\itshape #1.}}

\newcommand*\circled[1]{\tikz[baseline=(char.base)]{
		\node[shape=circle,draw,inner sep=0pt] (char) {#1};}}

\newcommand{\maxspeeduppublic}{$9.5\%$\xspace}
\newcommand{\maxspeedupkpi}{$9.5\%$\xspace}
\newcommand{\maxspeedupyahoo}{$4.8\%$\xspace}

\newcommand{\maxspeedupprivate}{$28.3\%$\xspace}

\definecolor{transparentblue}{RGB}{240, 240, 255}
\definecolor{lightblue}{RGB}{200, 200, 255}

\setlist[itemize]{itemsep=0pt,topsep=1pt,leftmargin=0.4cm}
\setlist[enumerate]{itemsep=0pt,topsep=1pt,leftmargin=0.4cm}

\title{\Large \bf \sys{}: Agentic Time-Series Anomaly Detection with Autonomous Rule Generation via Large Language Models}

\begin{document}

\date{}

\author[1,2*]{Yile Gu}
\author[2]{Yifan Xiong}
\author[2]{Jonathan Mace}
\author[2]{Yuting Jiang}
\author[1,3]{Yigong Hu}
\author[1]{Baris Kasikci}
\author[2]{Peng Cheng}

\affil[1]{University of Washington}
\affil[2]{Microsoft Research}
\affil[3]{Boston University}

\maketitle
\def\thefootnote{*}
\footnotetext{This work was primarily done during an internship at Microsoft Research.}
\def\thefootnote{\arabic{footnote}}

\begin{abstract}

Observability in cloud infrastructure is critical for service providers, driving the widespread adoption of anomaly detection systems for monitoring metrics.
However, existing systems often struggle to simultaneously achieve explainability, reproducibility, and autonomy, which are three indispensable properties for production use.
We introduce \sys{}, an agentic system for detecting time-series anomalies in cloud infrastructure by leveraging large language models (LLMs).
\sys{} proposes to use explainable and reproducible anomaly rules as intermediate representation and employs LLMs to autonomously generate such rules.
The system will efficiently train error-free and accuracy-guaranteed anomaly rules through multiple collaborative agents and deploy the trained rules for low-cost online anomaly detection.
Through evaluation results, we demonstrate that \sys{} outperforms state-of-the-art methods, increasing $F_1$ scores by up to \maxspeeduppublic and \maxspeedupprivate on public anomaly detection datasets and an internal dataset collected from \ifdraft an industry cloud\else\msft{}\fi, respectively.

\end{abstract}

\section{Introduction}

Ensuring the reliability and availability of cloud services is a key challenge for service providers~\cite{RCACopilot, Gandalf, Deoxys, Narya, Acto}, as downtime or interruptions can severely impact both customer experience and business operations.
In December 2021, an unexpected surge in connectivity triggered by autoscaling led to a major outage in Amazon Web Services (AWS), causing disruptions to downstream services and impacting millions of users worldwide for more than 10 hours~\cite{aws-outage-2021}.

To minimize the negative consequences of service interruptions, early-detection of anomalies in metrics monitoring  is crucial, as such metrics provide real-time insights into the health and performance of cloud services.
Large companies often develop and deploy anomaly detection systems in production environments~\cite{SRCNN,FBDetect,borg}. 
These systems are tailored to address the scale, complexity, and unique requirements of their vast and dynamic infrastructures.
For instance, Google's Borg~\cite{borg} features robust monitoring tools that track task health and performance metrics, automatically restarting failed tasks and scaling up to tens of thousands of machines.

\begin{figure}[t]
    \centering
    \begin{subfigure}[b]{0.49\linewidth}
        \centering
        \includegraphics[width=\linewidth]{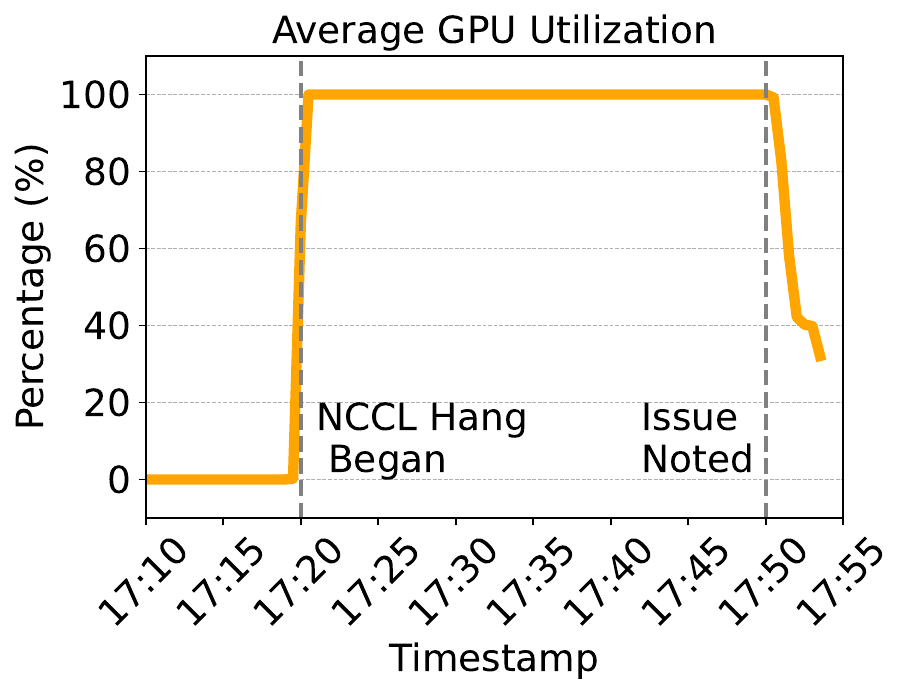}
    \end{subfigure}
    \hfill
    \begin{subfigure}[b]{0.49\linewidth}
        \centering
        \includegraphics[width=\linewidth]{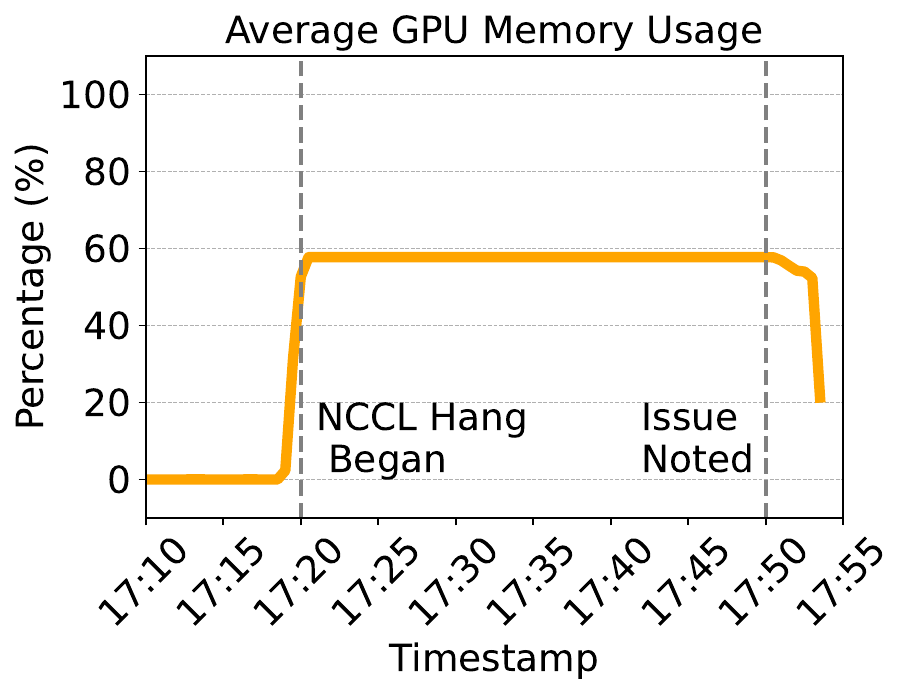}
    \end{subfigure}
    \caption{GPU utilization and memory usage metrics for a distributed model training on 256 A100 GPUs, where a hang issue occurred in NCCL since the job launched.}
    \label{fig:intro-nccl-hang}
\end{figure}

However, detecting anomalies in time with high accuracy is challenging due to the variety of anomalies.
For example, \Cref{fig:intro-nccl-hang} illustrates a real-world interruption, where distributed model training on 256 A100 GPUs encountered a network hang issue~\cite{ncclhang}.
At timestamp 17:20, the issue occurred, and GPUs started to busy wait---with high utilization and memory usage---for communication to resume and stalled the training process.
Although at first glance it seems normal that GPU utilization is saturated and GPU memory is effectively utilized,  in normal training process there should be variation for GPU utilization and GPU memory, which is not observed in the figure.
Monitors equipped with manually-written anomaly rules failed to detect the issue in time, which was only mitigated after human intervention at timestamp 17:50 by terminating training, resulting in significant waste of resources and time.
Such network hanging incidents have also been observed in multiple large companies\cite{sb, llama3, megascale}.
To improve the accuracy of the monitors, the engineers have to manually update them with new anomaly rules such as ``All GPUs are continuously running at 100\% utilization for periods exceeding 15 minutes''.

Prior work~\cite{Zhao2024TowardsRA,explainableDeepAD,explainablesurvey,automatedAD,automatedADYahoo}, as well as our own experience deploying anomaly detection systems for large-scale incident management at \msft{}, highlights three indispensable properties for designing such systems:

\begin{enumerate*}[label=(\roman*)]
    \item \textbf{Explainability}.
    No system is perfect, and false alarms are inevitable in anomaly detection systems.
    When alarms occur, on-call engineers (OCEs) must be able to understand the underlying reasons for these outcomes.
    An explainable anomaly detection system will enable OCEs to easily improve any inaccuracies.
    \item \textbf{Reproducibility}.
    The system should also produce consistent results for the same input metrics, ensuring that when an alarm is triggered in production, OCEs can reproduce it in a different environment and conduct further root cause analysis.
    A reproducible anomaly detection system will also eliminate non-deterministic alarms, avoiding wasted engineer effort.
    \item \textbf{Autonomy}.
    The system also needs to be frequently updated as data distributions shift when new metrics or dominant workloads are introduced~\cite{PerfAnalysisGmail,Gandalf}.
    An autonomous anomaly detection system can adapt to these changes in data distribution without any human intervention.
\end{enumerate*}

Prior work on time-series anomaly detection can be broadly categorized into three directions, yet none of these methods simultaneously address explainability, reproducibility, and autonomy.
\textit{Conventional deep learning-based methods}~\cite{AnomalyTransformer, AR, Donut, FCVAE, Informer, LSTMAD, SPOT, SRCNN, TFAD, TranAD} often lacks explainability since they generate anomaly labels directly from input data.
To improve model accuracy, engineers typically tune hyperparameters or model architectures, which makes these methods have only partial autonomy. 
\textit{LLM-based methods}~\cite{LLMTime,LLMAD,LLM4TSAD?,SigLLM} enhance explainability and autonomy by enabling OCEs to prompt with anomaly descriptions and providing anomaly labels along with explanations for why the data is classified as anomalous.
However, due to the inherent non-determinism of LLMs~\cite{song2024nondeterminism, ouyang2024nondeterminism}, these methods suffer from a lack of reproducibility and often produce inconsistent results when the same data is input across multiple trials.
Consequently, \textit{rule-based methods}~\cite{ServiceLab,FBDetect,Resin} are widely used in industry for time-series anomaly detection to achieve explainability and reproducibility.
These methods use anomaly detection rules which are easy for developers to understand, as the exact monitor logic is explicitly outlined and can be understood both prior to deploying the monitor and after the monitor has been triggered.
However, current rule generation and threshold tuning heavily rely on manual efforts, thereby lacking autonomy.
As demonstrated by the example in \Cref{fig:intro-nccl-hang}, this results in cases where monitors haven't been configured correctly, due to a lack of developer resources or human error in developing the rules.

This paper explores how to simultaneously achieve explainability, reproducibility, and autonomy in anomaly detection systems.
We observe that structured detection rules serve as an effective intermediate representation for such systems.
The rule-based method suggests that the detection rules can be written in the format of executable code, which is both reproducible and explainable.
On the other hand, LLMs can be leveraged to make rule generation autonomous for time-series anomaly detection.
LLMs have shown promising results in time-series tasks~\cite{Time-LLM,LLMTime,lagllama}, demonstrating a strong understanding of data patterns.
Moreover, LLMs can generate executable code for various tasks~\cite{codellama,codex,llmfuzzer}.
These capabilities make LLMs an ideal candidate for autonomously generating rules.

A key insight of our work is that integrating LLM-generated rules with classical rule-based methods bridges the gap between autonomy and explainability while retaining reproducibility.
Unlike existing LLM-based methods~\cite{LLMTime,LLMAD,LLM4TSAD?,SigLLM} that leverage LLMs at runtime detection, which often suffer from randomness and lack reproducibility, we use LLMs to identify and codify the anomaly rules in the training phase.
These explainable and reproducible rules are then deployed in runtime to detect anomalies.

Nevertheless, employing LLMs to generate anomaly detection rules and implement in executable code presents unique challenges.
First, LLMs can produce code or rules that have syntax errors or are inaccurate due to a misunderstanding of data patterns.
It is challenging to address the syntactic and accuracy issues.
Second, it is hard to guarantee that the anomaly rules autonomously generated by LLMs have better accuracy than existing production anomaly detection systems, which have been well-tuned over years of use.
Finally, due to the inherent randomness in LLM behavior, producing accurate anomaly detection rules with a limited number of trials (as LLM generation is expensive) remains challenging.

To address these challenges, we propose \sys{}, a time-series anomaly detection system that autonomously generates rules using LLMs.
First, \sys{} employs an agent-based pipeline with feedback loops to iteratively correct anomaly detection rules and improve accuracy.
In each iteration, multiple agents collaborate to propose, validate, fix, and refine the rules, reducing both syntax errors and improving accuracy.
Second, to further guarantee better accuracy, \sys{} aggregates
the predictions of its anomaly detection rules with the predictions of existing anomaly detection systems.
During training, \sys{} primarily learns data patterns from the incorrect samples identified by an existing anomaly detector.
During runtime inference, \sys{} uses an aggregation algorithm to merge the predictions from both the rules and the existing anomaly detectors to generate the final anomaly prediction.
Finally, to improve the efficiency of generatingaccurate anomaly detection rules, \sys{} simultaneously proposes a set of $n$ rule candidates in each iteration and selects the best $k$ rules for further refinement in the next iteration.

We evaluate \sys{} on two widely used public time-series anomaly detection datasets, KPI~\cite{kpidataset} and Yahoo~\cite{yahoodataset}, as well as an internal dataset collected from \msft{}.
Compared to the best baselines, our proposed system improves the average $F_1$ score by \maxspeedupkpi and \maxspeedupyahoo on the KPI~\cite{kpidataset} and Yahoo~\cite{yahoodataset} datasets, respectively.
It also achieves up to a significant \maxspeedupprivate $F_1$ score improvement on our internal dataset.
Besides, \sys{} speeds up inference by $3.0\times$, $34.3\times$, and $1.5\times$ on the KPI, Yahoo, and Internal datasets, respectively.

In summary, our contributions are as follows:
\begin{itemize}
    \item We show that current state-of-the-art time-series anomaly detection systems fall short in simultaneously achieving explainability, reproducibility, and autonomy.
    \item We observe that LLMs could be employed to autonomously generate explainable and reproducible rules for anomaly detection.
    \item We propose \sys{}, a time-series anomaly detection system that autonomously trains and deploys anomaly detection rules through an LLM-based agentic pipeline.
    \item We evaluate \sys{} on public and internal datasets, demonstrating its effectiveness and efficiency compared with state-of-the-art methods.
\end{itemize}

\section{Motivation and Challenges}

\subsection{Why Existing Methods Fall Short?}
\label{subsec:motivation-tsad}

\begin{figure}[t]
    \centering
    \begin{subfigure}[b]{0.48\linewidth}
        \centering
        \includegraphics[width=\linewidth]{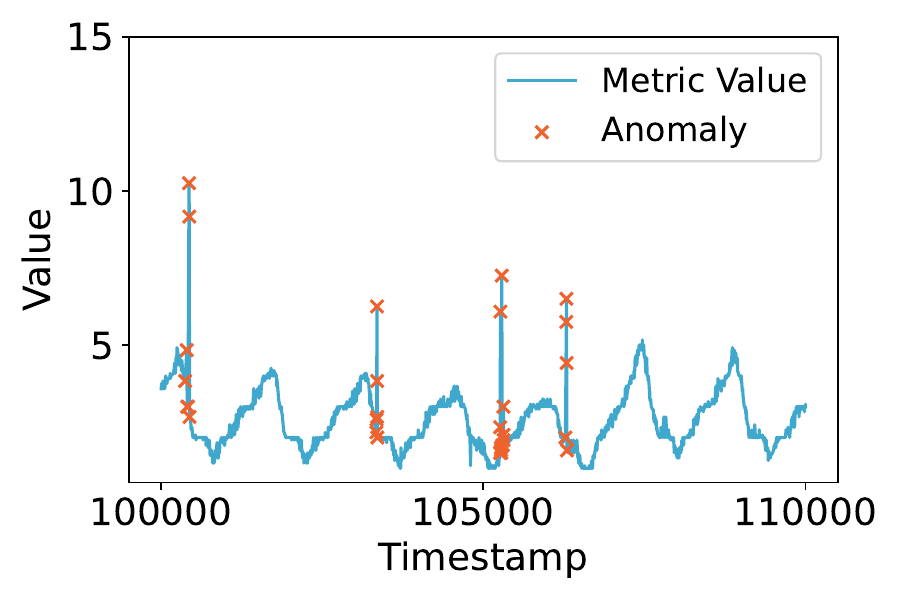}
        \caption{Metric \texttt{da403}}
        \label{fig:motivation-da403}
    \end{subfigure}
    \hfill
    \begin{subfigure}[b]{0.48\linewidth}
        \centering
        \includegraphics[width=\linewidth]{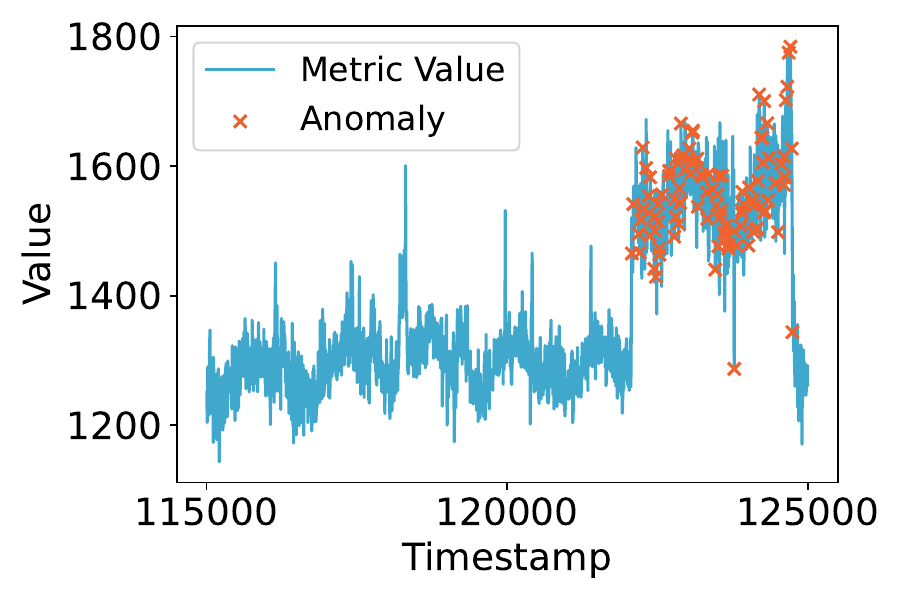}
        \caption{Metric \texttt{1c35d}}
        \label{fig:motivation-1c35d}
    \end{subfigure}
    \caption{Two example metrics from the KPI dataset, where the blue line shows the metric data and the orange crosses represents the anomalies.}
    \label{fig:motivation}
\end{figure}

\textbf{Background.}
In cloud infrastructure, various metrics such as CPU usage, memory consumption, network latency, disk I/O, and others are periodically collected and continuously monitored to ensure the health of cloud services~\cite{zhang2024surveyaiops, remil2024survelaiops, cheng2023surveyaiops}.
Anomalies, which represent deviations in key operational metrics that significantly diverge from normal patterns, can be caused by various factors in production, including hardware failures, software bugs, resource contention, and so on~\cite{FBDetect,llama3,megascale,vprof,violet}.
In practice, there is a variety of anomaly patterns, and each application or each metric in the cloud infrastructure may exhibit different behaviors.
\Cref{fig:motivation} shows two example metrics from the KPI dataset~\cite{kpidataset}.
The anomalies in~\Cref{fig:motivation-da403} are characterized by multiple sudden spikes in the metric values, while the anomalies in~\Cref{fig:motivation-1c35d} are represented by a persistent shift-up.

\begin{table}[t]
\centering
\caption{$F_1$ scores of different methods on example metrics.}
\begin{tabular}{r@{\hspace{.4em}}lccc}
\toprule
\multicolumn{2}{c}{\textbf{Method}} & \textbf{\texttt{da403}} & \textbf{\texttt{1c35d}} \\
\midrule
FCVAE & (best setting) & 0.97 & 0.80 \\
\midrule
\multirow{3}{*}{Manual Rule} & (\stt{threshold=1}) & 0.05 & 0.01 \\
 & (\stt{threshold=3}) & \textbf{0.99} & 0.17 \\
 & (\stt{threshold=5}) & 0.94 & 0.43 \\
\midrule
\multirow{2}{*}{LLM Rule} & (for \texttt{da403}) & \textbf{0.99} & N/A \\
 & (for \texttt{1c35d}) & N/A & \textbf{0.91} \\
\bottomrule
\end{tabular}
\label{tab:motivation-f1-scores}
\end{table}

\begin{figure}[t]
    \centering
    \begin{subfigure}[t]{\linewidth}
        \centering
        \begin{lstlisting}[language=Python,style=customnew]
def rule(sample: np.ndarray, threshold: float) 
    -> np.ndarray:
    # Get values and create labels
    values = sample[:, 0]
    labels = np.zeros(sample.shape[0])
    # Calculate mean and standard deviation
    mean_val = np.mean(values)
    std_val = np.std(values)
    # If a value is more than 
    # threshold * standard deviations away 
    # from the mean, it is considered abnormal
    labels[np.abs(values - mean_val) > 
        threshold * std_val] = 1
    return labels\end{lstlisting}
    \end{subfigure}
    \caption{An example rule written by human and implemented in Python, modified from~\cite{Resin}.}
    \label{lst:manual_rule}
\end{figure}

\textbf{Rule-Based Methods.}
In monitoring systems, it is common practice for engineers to write rules to detect anomalies for monitoring metrics~\cite{ServiceLab,FBDetect,Resin}.
These rules are typically based on domain knowledge and the engineers' experience with the metrics.
\Cref{lst:manual_rule} demonstrates an example anomaly detection rule implemented in Python to detect anomalies in time-series data, adapted from the rule used in~\cite{Resin}.
The \stt{rule} function takes a time-series \stt{sample} and a \stt{threshold} as input, and returns a sequence of \stt{labels} indicating whether each data point is abnormal or not.

\Cref{tab:motivation-f1-scores} compares the $F_1$ scores of the FCVAE model~\cite{FCVAE}, a state-of-the-art DL-based model, and the manual rule with different thresholds on the two example metrics shown in~\Cref{fig:motivation}.
We observe that the result of the manual rule is highly sensitive to the specific characteristics of the anomalies in each metric, thereby which may be tackled using specific thresholds yielding a high $F_1$ score. While the manual rule performs similarly to the FCVAE model on the \texttt{da403} metric, it consistently underperforms on the \texttt{1c35d} metric across different thresholds.
This discrepancy is caused by different anomaly patterns in two metrics, highlighting that no single `one-size-fits-all' rule can effectively handle all types of anomalies.
When new anomalies or metrics emerge in cloud services, engineers must manually design or adjust anomaly detection rules, which is time-consuming and requires significant expertise~\cite{FBDetect}.

\begin{figure}[t]
    \centering
    \begin{subfigure}[b]{0.48\linewidth}
        \centering
        \includegraphics[width=\linewidth]{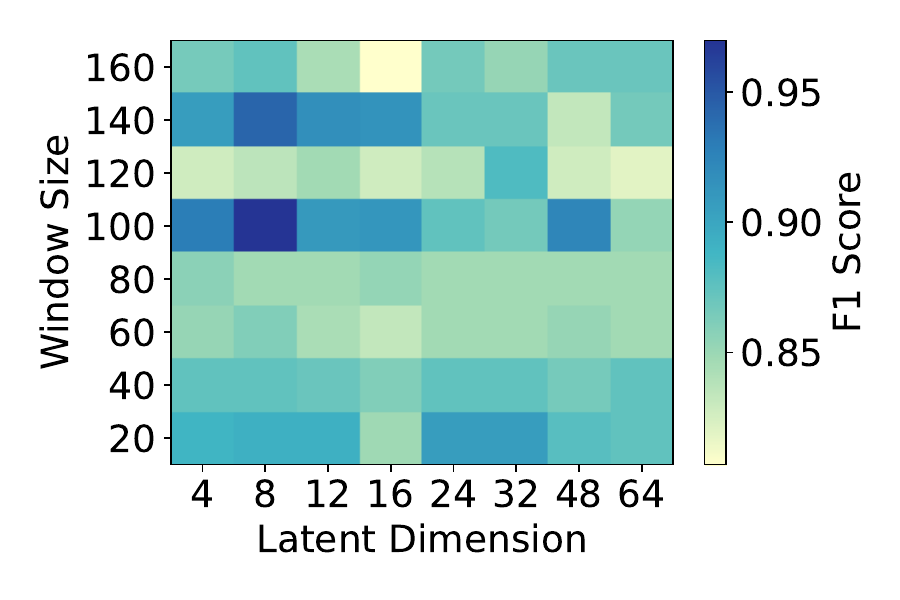}
        \caption{Grid search for FCVAE}
        \label{fig:motivation-hyperparameter-search}
    \end{subfigure}
    \hfill
    \begin{subfigure}[b]{0.48\linewidth}
        \centering
        \includegraphics[width=\linewidth]{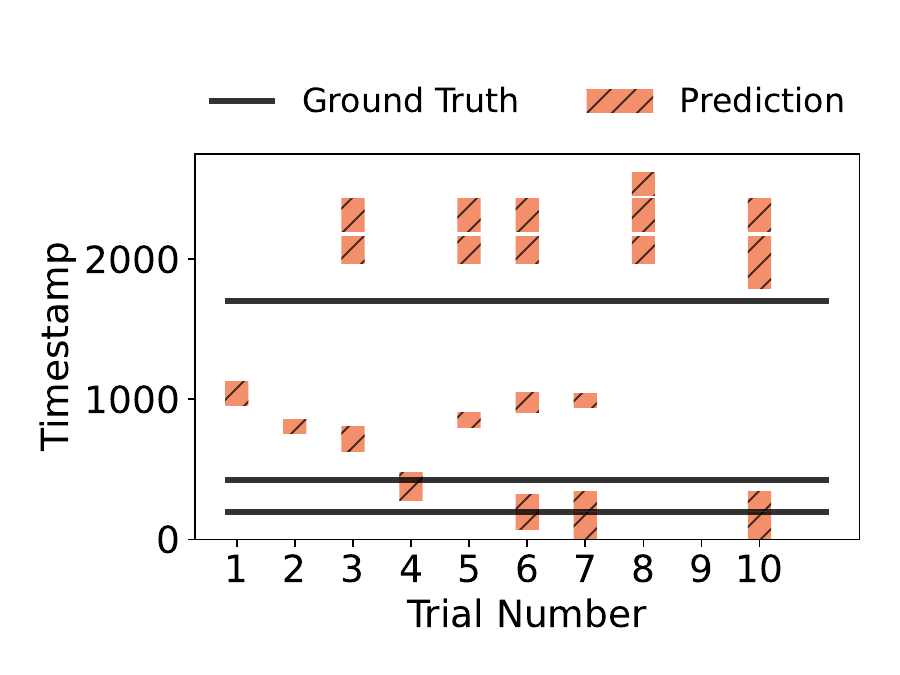}
        \caption{SigLLM results over 10 trials} %
        \label{fig:motivation-variability}
    \end{subfigure}
    \caption{Analysis of prior methods on the KPI dataset metric \texttt{da403}.}
    \label{fig:motivation-prior-approaches}
\vspace{-1em}
\end{figure}

\textbf{Conventional DL-Based Methods.}
Deep learning (DL) models have shown promising results in time-series anomaly detection by learning data labels to classify anomalies~\cite{AnomalyTransformer, AR, Donut, FCVAE, Informer, LSTMAD, SPOT, SRCNN, TFAD, TranAD}.
However, the lack of explainability in these models poses a significant challenge for engineers in practice.
These DL models contain many hyperparameters, which are difficult to interpret in terms of their impact on anomaly detection performance.
As a result, engineers often need to perform an exhaustive grid search across a wide range of hyperparameters to find the optimal configuration, which can be both time-consuming and computationally expensive.

\Cref{fig:motivation-hyperparameter-search} shows the grid search results for the FCVAE model~\cite{FCVAE} on the \texttt{da403} metric in the KPI dataset, considering two hyperparameters: the latent dimension and the window size.
The latent dimension defines the dimension of the hidden space where the time-series input is projected, while the window size determines the number of data points the model uses for prediction.
We observe that the model $F_1$ score varies significantly across different hyperparameter configurations, indicating the difficulty of finding the optimal hyperparameters for the model.
Although a larger latent dimension incorporates more information in the hidden space, it does not always result in better model performance.
For the window size, setting it to 100 or 140 performs significantly better than a setting of 120, yet there is no clear intuition behind this performance difference.

\textbf{LLM-Based Methods.}
Recent advancements in large language models (LLMs) have expanded their application in time-series anomaly detection~\cite{LLMTime,LLMAD,LLM4TSAD?,SigLLM}, where they accept data from prompts to predict anomalies and optionally generate natural language explanations, thereby enhancing result explainability.
However, due to the stochastic nature of LLMs, their outputs can exhibit high variance across different trials, sacrificing reproducibility.

\Cref{fig:motivation-variability} shows the $F_1$ scores of the SigLLM system~\cite{SigLLM} over ten trials on the \texttt{da403} metric.
We observe that the output labels of the LLMs can vary significantly in terms of the number and locations of the anomalies detected, with no two trials producing the same predictions.
Four out of ten trials (No. 4, 6, 7, and 10) are able to detect one anomaly partially, while there exists three ground-truth anomalies in the metric.
If a majority vote is performed to produce a consistent final result from these trials~\cite{selfconsistency}, none of the anomalies will be correctly detected.

\begin{figure}[t]
    \centering
    \begin{subfigure}[b]{.98\linewidth}
        \centering
        \begin{lstlisting}[language=Python,style=customnew]
values = sample[:, 0]
labels = np.zeros(sample.shape[0])
# Calculate z-scores
z_scores = np.abs(stats.zscore(values))
# Abnormal Rule 1: If z-score is greater
# than 3, the data point is abnormal
labels[z_scores > 3] = 1\end{lstlisting}
        \vspace{-.5em}
        \caption{Metric \texttt{da403}}
        \label{fig:rule_LLM_da403}
    \end{subfigure}
    \hfill
    \begin{subfigure}[b]{.98\linewidth}
        \centering
        \begin{lstlisting}[language=Python,style=customnew]
for i in range(window_small, len(values)):
    last_N = values[i-window_small:i]
    last_M = values[i-window_large:i]
    avg_N = np.mean(last_N)
    avg_M = np.mean(last_M)
    # Abnormal Rule 1: A value differs from 
    # the mean of last N values by over 20%
    if values[i] > 1.2 * avg_N or \
        values[i] < 0.8 * avg_N:
        labels[i] = 1
    # Abnormal Rule 2: The mean of last N 
    # values differs from the mean of last M 
    # values by more than 20%
    elif avg_N > 1.2 * avg_M or \
        avg_N < 0.8 * avg_M:
        labels[i-window_small:i] = 1\end{lstlisting}
        \vspace{-.5em}
        \caption{Metric \texttt{1c35d}}
        \label{fig:rule_LLM_1c35d}
    \end{subfigure}
    \caption{Anomaly rules and code generated by LLM.}
    \label{fig:llm_rules}
\end{figure}

\subsection{How LLMs can Help in Rule Generation?}

Although anomaly detection rules are not autonomously generated, they are both explainable and reproducible.
For instance, the manual rule in~\Cref{tab:motivation-f1-scores} provides an explainable parameter \stt{threshold}, which allows engineers to tune the sensitivity of the rule to deviations from the mean.
Since the threshold is fixed during deployment and the rule's logic is deterministic, it also ensures the reproducibility of the detection results.

To address the lack of autonomy in rule generation, our intuition is to leverage LLMs to create and tune anomaly detection rules for time-series data, and to implement them in executable code, mimicking the rule development process of engineers.
Although LLMs have been used in time-series anomaly detection, prior work~\cite{LLMTime,LLMAD,LLM4TSAD?,SigLLM} only focuses on involving LLMs during deployment time to directly predict anomalies from the time-series data.
Instead, we propose to use LLMs in a training phase before the deployment to generate anomaly detection rules from time-series data collected and labeled offline.
The rules are then deployed in production to monitor the metrics, where LLMs are not involved in the real-time detection process.

\subsection{Opportunities and Challenges in Applying LLMs}

\Cref{fig:llm_rules} shows the promising results of anomaly rules and code generated by the LLM for the two example metrics.
We prompt the LLM with time-series data and ask it to first propose an anomaly detection rule in natural language and then write a Python implementation.
For the \texttt{da403} metric, the LLM generates a rule functionally equivalent to the manual rule, where data points are flagged as anomalies if their z-scores are greater than 3.
For the \texttt{1c35d} metric, the LLM generates a more complex rule that detects both spikes (\stt{Abnormal Rule 1}) and level shifts (\stt{Abnormal Rule 2}) in the metric values.
Both LLM-generated rules achieve similar or better $F_1$ scores compared to the manual rules, as shown in ~\Cref{tab:motivation-f1-scores}, demonstrating the potential of LLMs in generating effective rules for time-series anomaly detection while simultaneously satisfying explainability, reproducibility, and autonomy.

Although the initial results are promising, there exist several unique challenges in applying LLMs to anomaly detection rule generation.

\textbf{Correctness and Accuracy Issues.}
The anomaly detection rules are implemented in code.
Despite being pretrained on billions of lines of public source code~\cite{codex}, LLMs may still generate rule implementations that contain syntax errors.
Additionally, misunderstandings in the patterns of time-series data can lead to inaccuracy in both the description of the anomaly detection rules and their implementation.

For example, when using LLMs to generate anomaly detection rules 50 times for each metric in the KPI dataset, we observe an overall rate of 4.8\% syntax errors in the code implementation.
In addition, for the rules that are correct, the average $F_1$ score across all metrics is only 0.129, while the best deep learning-based method achieves a score of 0.819.

\textbf{No Accuracy Guarantee.}
A mature anomaly detection system in production will likely have well-tuned models for each metric over time.
Even if correctness and accuracy issues in rule generation are addressed, there is no guarantee that LLM-generated rules will always outperform existing mature anomaly detection systems.
For instance, compared to the best existing model in the KPI dataset, we observe that the best LLM-generated rules have  $F_1$ score regression with respect to 3 metrics out of 19 metrics.

\textbf{Low Efficiency.}
LLMs can generate a diverse range of responses even under identical inputs.
As a result, the anomaly detection system will need to invoke the LLM for multiple trials repeatedly, with each trial reflecting a different interpretation of the input data by the LLM, leading to variability in the accuracy of the generated rules.
Achieving accurate rules with a small number of trials is challenging due to the inherent variability in LLM outputs.
For instance, the LLM takes an average of 113 minutes to converge to an accurate rule on the KPI dataset.

\section{System Design}

\subsection{Overview}
\label{subsec:design-overall}

\begin{figure}[t]
    \centering
    \includegraphics[width=\linewidth]{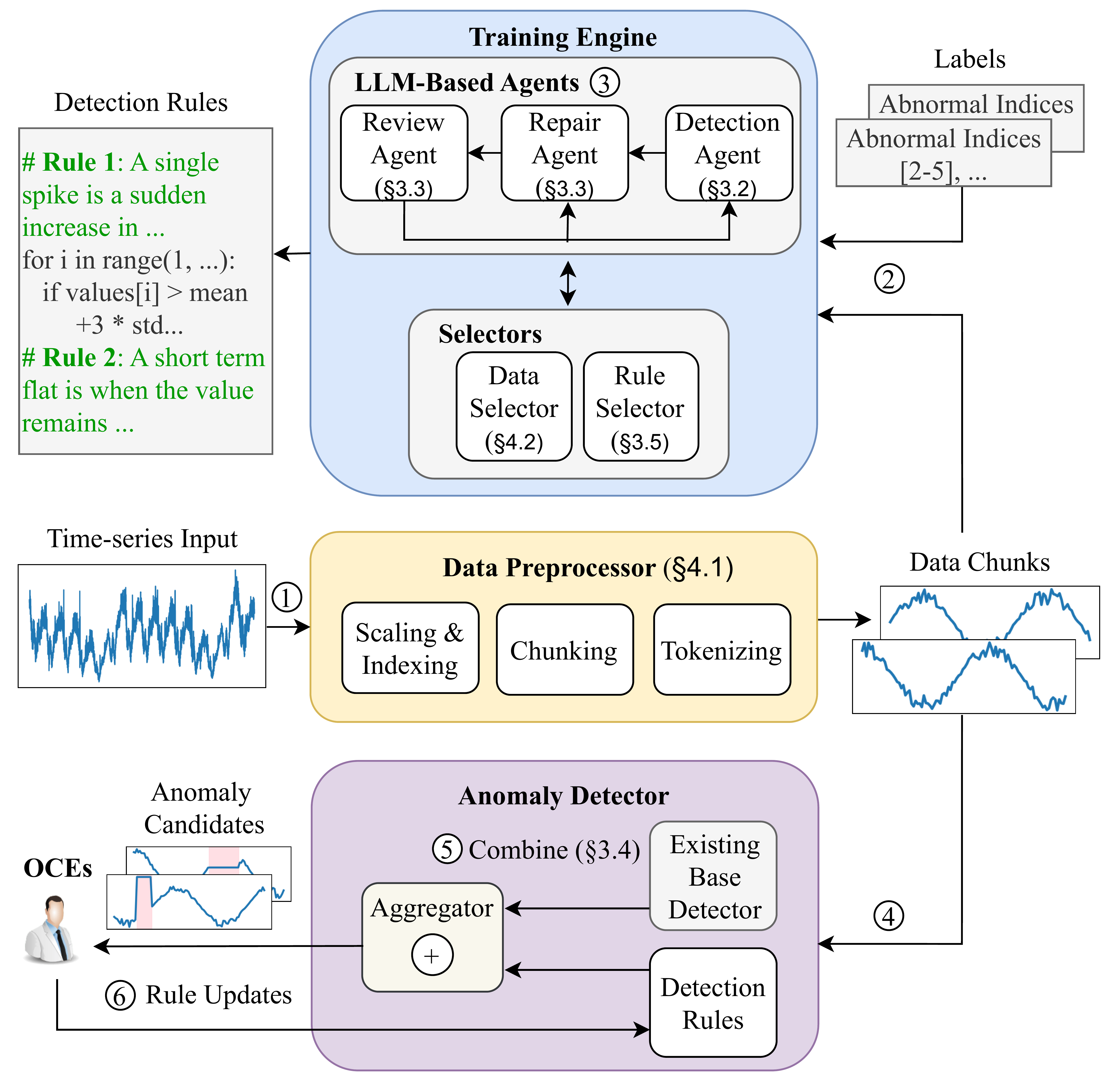}
    \caption{The overall design of \sys{}.}
    \label{fig:design}
\vspace{-0.3em}
\end{figure}

As shown in \Cref{fig:design}, \sys{} detects anomalies in time-series data using a three-stage approach: data preprocessing, rule training, and deployment.

\begin{enumerate}
    \item \textbf{Data Preprocessing.}
    Upon receiving a time-series input, \circled{1} the Data Preprocessor first chunks the input into smaller segments and applies data preprocessing techniques such as scaling and indexing.
    \item \textbf{Rule Training.}
    During the rule training stage, \circled{2} the data chunks and corresponding ground-truth labels are fed into the Training Engine, which iteratively trains detection rules based on the preprocessed data chunks and labels using an agentic pipeline.
    The pipeline consists of three agents: the Detection Agent, the Repair Agent, and the Review Agent.
    \circled{3} In each iteration, the Detection Agent first proposes a set of detection rules based on the input data.
    The Repair Agent then checks the proposed rules for syntax errors and corrects any issues.
    Next, the Review Agent evaluates the accuracy of the proposed rules using validation data.
    If any issues are detected, the rules will be sent back to the Repair Agent; otherwise, they will be fed back to the Detection Agent to incorporate new rules.
    This iterative loop continues to improve accuracy monotonically.
    \item \textbf{Deployment.}
    During deployment, \circled{4} the Anomaly Detector receives processed data chunks from the runtime time-series data.  
    \circled{5} An Aggregator combines the outputs from a base detector in the existing system and the anomaly detection rules to identify anomalies in the input data chunks.
    If an anomaly is detected, the Anomaly Detector will report an incident, allowing on-call engineers to perform root cause analysis and incident mitigation.
    \circled{6} Engineers can also update the detection rules based on incident feedback to improve detection accuracy.
\end{enumerate}

\begin{figure}[t]
  \begin{lstlisting}[language=Python,style=customblackwhite]
  # import necessary libraries for your code
  def inference(sample: np.ndarray) 
      -> np.ndarray:
      labels = np.zeros(len(sample))
      # Your comment to describe 
      # how normal data behave
      # Normal Rule 1
      # Normal Rule 2
      # Your code to detect 
      # if the given sample is abnormal
      # Abnormal Rule 1
      if ...
      # Abnormal Rule 2
      if ...
      # return labels as a 1d numpy array
      return labels\end{lstlisting}
  \caption{Code Template for the Detection Agent in \sys{}.}
  \label{fig:design-detection-agent-code-template}
\end{figure}

\subsection{Autonomous Rules Generation}
\label{subsec:design-detection-agent}

Detection rules must be in a format suitable for LLM generation, easily understandable by engineers, and generalizable across different types of anomalies.
\sys{} chooses Python as the implementation language for anomaly detection rules.
Python is the dominant programming language in datasets used to pretrain LLMs~\cite{codex}, and it is commonly used as the target language when finetuning LLMs for code-related tasks~\cite{codellama}.
Furthermore, as a Turing-complete language, it can perform any computation required for detecting anomalies in time-series data.

\Cref{fig:design-detection-agent-code-template} shows the code template for the Detection Agent in \sys{}.
The Detection Agent is tasked with writing a Python function, \stt{inference(sample: np.ndarray) -> np.ndarray}, to output prediction labels based on the input sample.
We prompt the Detection Agent to first describe the abnormal rules in the form of comments, and then write the corresponding Python code that implements these abnormal rules.
This step-by-step process generates commented anomaly detection rules that are easily understandable and verifiable by engineers.
Additionally, the Detection Agent is asked to describe the behavior of normal data in the comments and ensure that the implementation of the abnormal rules does not conflict with the normal rules.
The complete prompt used for the Detection Agent could be found in~\Cref{sec:appendix-prompts}.

\subsection{Correctness and Accuracy Improvement via Feedback Loops}
\label{subsec:design-repair-review-agent}

Despite clear instructions in the prompt, LLMs may still generate anomaly detection rules with syntax errors or inaccuracies, making it difficult to ensure the correctness and accuracy.
Inspired by backpropagation in deep learning training~\cite{sgd, efficientbackprop}, \sys{} introduces an iterative feedback loop involving two agents, the Repair Agent and the Review Agent, to train rules in an iterative manner.
The Repair Agent corrects syntax errors in the generated anomaly detection rules, while the Review Agent verifies their accuracy on a validation set.
In each iteration, the two agents collaborate together to ensure that the anomaly detection rules are syntactically correct and that their performance on the validation set is at least as good as the previous iteration.

\textbf{Repair Agent.}
Upon receiving anomlay detection rules from the Detection Agent, the Repair Agent first checks for syntax errors by executing the rules on dummy data that mimics the format of the input data.
If syntax errors are detected, \sys{} provides error messages and call stack information to the Repair Agent, which then proposes a corrected version of the rules using a new prompt and re-checks them until all syntax errors are resolved.

\textbf{Review Agent.}
The Review Agent evaluates the accuracy of the anomaly detection rules on the validation data.
If the accuracy of the current anomaly detection rules is worse than the previous iteration, \sys{} provides a comparison of the accuracy metrics and the code differences between the two versions to the Review Agent.
\sys{} also provides data samples where the rule from the previous iteration labels correctly, while the new rule labels incorrectly.
The Review Agent then proposes an improved version of the rules based on these observations and re-evaluates the accuracy until there is no accuracy regression in current iteration.
If the new anomaly detection rules contain syntax errors, they will be sent back to the Repair Agent for further correction.
In practice, this process will not block training infinitely, as the Review Agent can revert the code changes from the current iteration, restoring accuracy to the previous iteration’s level.

\subsection{Accuracy Guarantee via Model Fusion}
\label{subsec:design-combined-mode}

Due to the lack of domain knowledge, LLMs may generate anomaly detection rules that underperform compared to existing anomaly detectors that have been well-tuned over time in production.
Instead of directly using the LLM-generated rules, \sys{} proposes a model fusion approach that combines these rules with the well-established anomaly detectors deployed in production, ensuring an accuracy guarantee.

During the training stage, \sys{} separately trains two sets of anomaly detection rules: one from false negatives and the other from false positives outputted by the existing base detector.
\sys{} defines false negatives to be ground-truth anomaly samples that are mis-classified by the base detector to be normal.
Similarly, false positives are normal samples that are mis-classified by the base detector to be abnormal.
To train the set of anomaly detection rules from false negative samples, \sys{} first performs inference on the training data using the existing base detector and identifies false negatives by comparing the predictions of the base detector with ground-truth labels.
It then trains anomaly detection rules by feeding the false negative samples and their corresponding ground-truth labels into the Training Engine.
To prevent overfitting to the false negatives, \sys{} also provides sampled true negative data to the Training Engine, and the Detection Agent is instructed to generate rules that detect anomalies in the false negative samples while excluding the true negative samples.
In each iteration, the validation accuracy of the anomaly detection rules is evaluated by combining the outputs of the rules and the existing base detector on the validation set using the Aggregator.
\sys{} trains detection rules for false positives in a similar manner.

During the deployment stage, the Anomaly Detector receives processed data chunks from the runtime time-series data.
The Aggregator then combines the output labels from the existing base detector with the anomaly detection rules for false negatives and false positives to identify anomalies in the input time-series data chunks.

\begin{algorithm}[t]
\caption{Anomaly Prediction Aggregation Algorithm}
\label{alg:aggregation}
\begin{algorithmic}[1]
\Require predicted labels from base detector $L_{base}$, from false positive rules $L_{fp}$, and from false negative rules $L_{fn}$
\Ensure final predicted labels $L_{agg}$
\Function{Aggregation}{$L_{base},~L_{fp},~L_{fn}$}
  \State $L_{agg} \gets L_{base}$
  \For{$t \gets 1$ to $length(L_{base})$}
    \If{$L_{base}[t] = \texttt{normal}$ \textbf{and} $L_{fn}[t] = \texttt{abnormal}$}
      \State $L_{agg}[t] \gets \texttt{abnormal}$
    \EndIf
    \If{$L_{base}[t] = \texttt{abnormal}$ \textbf{and} $L_{fp}[t] = \texttt{normal}$}
      \State $L_{agg}[t] \gets \texttt{normal}$
    \EndIf
  \EndFor
  \State \Return $L_{agg}$
\EndFunction
\end{algorithmic}
\end{algorithm}

\textbf{Aggregator.}
\Cref{alg:aggregation} presents the aggregation algorithm for the Aggregator in \sys{}.
The anomaly detection rule for false negatives effectively corrects the normal labels from the base detector but does not address the abnormal labels.
Similarly, the anomaly detection rule for false positives corrects the abnormal labels from the base detector but does not affect the normal labels.
This is because a set of anomaly detection rules only look at either false negative or false positive examples at a time during the training, which makes the anomaly detection an one-class classification task~\cite{deeponeclass}.

\subsection{Efficiency Enhancement}
\label{subsec:design-rule-selector}

Due to the stochastic nature of the autoregressive process in LLMs, they may not generate optimal anomaly detection rules within a small number of trials.
Inspired by the beam search algorithm~\cite{beamsearch}, \sys{} employs a top-$k$ selection strategy to identify the best rules and early terminates the inaccurate rules during training.
In each iteration, the Detection Agent uses the same input to propose $n$ detection rules, which are passed to both the Repair Agent and the Review Agent.
Once the Review Agent verifies that there is no accuracy regression in the generated $n$ rules, the Rule Selector selects the top-$k$ rules based on user-defined criteria.
Currently, \sys{} chooses validation accuracy as the criterion, selecting the $k$ rules with the highest accuracy on the validation set.
The Rule Selector can also be extended to incorporate additional criteria, such as the inference time of the anomaly detection rules.

\section{Implementation}

\subsection{Data Preprocessing}
\label{subsec:design-data-preprocessor}

Since time-series data are continuous and unbounded, they must first be chunked into smaller segments for processing by the Training Engine.
Meanwhile, the Data Preprocessor also applies scaling, indexing, and tokenization-specific preprocessing, which are essential for LLMs to learn patterns in the data.

\textbf{Scaling and Indexing.}
Time-series inputs may have arbitrary precisions, which limits the amount of data that can fit within the context window of LLMs.
To address this, the Data Preprocessor scales the input data to a specified significant figure.
Similarly, timestamps, often represented by long numeric values, are not informative to LLMs.
Therefore, the Data Preprocessor removes the timestamp and re-indexes the data based on the order of the data points.
\sys{} assumes that data points are collected at fixed time intervals.

\textbf{Chunking.}
Selecting an appropriate chunk size is crucial for training anomaly detection rules.
A chunk size that is too small may lack sufficient contextual information for the LLMs to learn the data patterns, while a chunk size that is too large may exceed the context window size of LLMs or result in overly general rules.
To determine an appropriate chunk size, \sys{} performs calibration on a metric of a given dataset.
\sys{} uses only the Detection Agent to repeatedly propose anomaly detection rules on the given metric, and selects the chunk size that results in the highest $F_1$ score on the training set.

\textbf{Tokenizer-Specific Preprocessing.}
Recent studies have shown that different tokenizers handle numerical values in different ways~\cite{SigLLM,nogueira2021investigating,LLMTime}, which impacts how LLMs interpret the data.
For the same numerical value, tokenizers may segment the value into chunks that do not align with its individual digits.
To address this issue, the Data Preprocessor applies tokenizer-specific preprocessing to the data based on the type of LLM used in the Training Engine.
For GPT-based models, the Data Preprocessor adds space between each digit in a time-series value, similar to techniques used by prior work~\cite{SigLLM}. 

\subsection{Data Selection}

\textbf{Automatic Metrics Selection.} Given a time-series dataset, the Data Selector controls how the dataset is partitioned and which partition to use during the training process.
The Data Selector supports two dataset modes: \textit{one-for-one} and \textit{one-for-all}.
The one-for-one mode is suitable for datasets with a large number of continuous metrics, as it allows the Training Engine to focus on learning the patterns of each metric individually.
The one-for-all mode is useful when the user wants to train a set of anomaly detection rules that generalize across multiple metrics.
\sys{} first chunks each metric in the dataset using the chunk size determined by the Data Preprocessor.
For metrics with at least 10 data chunks (i.e., where at least 10 training iterations can be run), \sys{} uses one-for-one mode to train these metrics.
For the rest metrics, \sys{} uses the one-for-all mode.
Additionally, \sys{} allows users to directly specify the dataset modes.

In the one-for-one mode, the Data Selector partitions the dataset based on the number of continuous time-series metrics.
In each training trial, the Data Selector selects one continuous metric from the dataset to train a set of anomaly detection rules.

In the one-for-all mode, the Data Selector partitions the dataset based on a group identifier provided by the user.
Each group contains a set of continuous time-series metrics.
In each training trial, the Data Selector selects one group from the dataset to train a set of anomaly detection rules.
This mode also serves as a data augmentation technique when the length of each continuous metric is short.

\textbf{Contrastive Example Retrieval.}
When training anomaly detection rules from false negative or false positive examples of existing anomaly detection models, the Data Selector needs ensure that the contrasitive examples retrieved (i.e. true negatives or true positives) are from a similar distribution as the false negative or false positive examples.
This is because selecting true negative or true positive examples from a different data distribution may lead to false assumptions about the data patterns, which can result in the generation of inaccurate anomaly detection rules.

In the \textit{one-for-one} mode, the Data Selector retrieves true negative or true positive examples randomly, since all data samples are from the same distribution.
In the \textit{one-for-all} mode, the Data Selector first calculates the mean and standard deviation of the false negative or false positive examples, then retrieves true negative or true positive examples sorted by their Euclidean distance to the mean and standard deviation of the false negative or false positive examples.
We believe more advanced retrieval techniques based on embedding similarity can be applied to further improve the retrieval quality and leave it as future work.

\section{Evaluation}

\subsection{Experiment Setup}

\begin{table}[t]
    \centering
    \caption{An example of different evaluation metrics in time-series anomaly detection.}
    \begin{tabular}{ccccccc}
    \multicolumn{7}{c}{\textbf{Ground-Truth}: \texttt{[0, 1, 1, 0, 1, 1, 1, 1, 0, 0]}} \\
    \multicolumn{7}{c}{\textbf{~~~~~Predictions}: \texttt{[1, 1, 0, 0, 0, 0, 0, 1, 1, 1]}} \\
    \toprule
    \textbf{Eval Method} & \textbf{TP} & \textbf{FP} & \textbf{FN} & \textbf{Pr} & \textbf{Re} & \textbf{$F_1$} \\ 
    \midrule
    Point-$F_1$    & 2 & 3 & 4 & 0.40 & 0.33 & 0.36 \\
    Point-$F_1$ PA & 6 & 3 & 0 & 0.66 & 1.00 & 0.80 \\
    Overlap-$F_1$  & 2 & 0 & 0 & 1.00 & 1.00 & 1.00 \\
    Event-$F_1$ PA & 2 & 3 & 0 & 0.40 & 1.00 & 0.57 \\
    \bottomrule
    \end{tabular}
    \label{tab:evaluation-metrics}
\end{table}

\textbf{Metrics.}
\sys{} uses the $F_1$ score as the primary evaluation metric when comparing against baselines.
The $F_1$ score is calculated as the harmonic mean of precision and recall.
Precision is the ratio of true positives (TP) to the sum of true positives and false positives (FP), while recall is the ratio of true positives to the sum of true positives and false negatives (FN).
Formally, the $F_1$ score is calculated as:
\begin{equation}
    F_1 = 2 \times \frac{\text{Precision} \times \text{Recall}}{\text{Precision} + \text{Recall}}
\end{equation}
where $\text{Precision} = \frac{\text{TP}}{\text{TP}~+~\text{FP}}$ and $\text{Recall} = \frac{\text{TP}}{\text{TP}~+~\text{FN}}$.

In time-series anomaly detection, defining positive and negative samples in the context of the application is crucial~\cite{si2024timeseriesbench}.
In cloud infrastructure monitoring, each anomaly is treated as an incident, which may span multiple data points across time.
For example, the ground-truth (GT) labels in~\Cref{tab:evaluation-metrics} contain two incidents: one spanning indices 1-2 and the other spanning indices 4–7.
The prediction labels correctly identify both incidents, but only partially overlap with the ground-truth indices.
As illustrated in~\Cref{tab:evaluation-metrics}, four different evaluation methods are used by prior work to compute the $F_1$ score for example predictions based on the ground-truth:
\begin{itemize}
    \item Point-Based $F_1$ Score (Point-$F_1$)~\cite{f1eval}: Treats each data point as an individual instance, often leading to a very low $F_1$ score.
    \item Point-Based $F_1$ Score with Point Adjustment (Point-$F_1$ PA)~\cite{TFAD}: Considers all points within a segment to be detected if at least one point in the segment is detected, often leading to a high $F_1$ score.
    \item Overlap $F_1$ Score (Overlap-$F_1$)~\cite{orion}: Treats each incident as a single instance but ignores false positives in the prediction, often resulting in an unreasonably high $F_1$ score.
    \item Event-Based $F_1$ Score with Point Adjustment (Event-$F_1$ PA)~\cite{eventf1pa}: Treats each incident as a single instance and penalizes precision by treating each data point outside the ground-truth incident as a false positive, leading to a more balanced $F_1$ score.
\end{itemize}
In \sys{}, we use Event-$F_1$ PA as the primary evaluation method for calculating the $F_1$ score metric.

\textbf{Baselines.}
We compare the proposed \sys{} against various state-of-the-art methods, including five DL-based methods from EasyTSAD framework~\cite{si2024timeseriesbench} and two LLM-based methods:

\begin{itemize}
    \item AnomalyTransformer~\cite{AnomalyTransformer}: An unsupervised model using the Anomaly-Attention mechanism to detect anomalies by exploiting differences in association patterns between normal and abnormal points.
    \item AutoRegression~\cite{AR}: A supervised model with several linear layers that transform input data into anomaly score logits with same length as the input.
    \item FCVAE~\cite{FCVAE}: An unsupervised model that integrates both global and local frequency features  to capture similar yet different periodic patterns and detailed trends in time series.
    \item LSTMAD~\cite{LSTMAD}: A supervised LSTM model that trains on normal data and detects anomalies using a statistical strategy based on the prediction error for observed data.
    \item TFAD~\cite{TFAD}: A supervised model that uses time-series decomposition to transform input data into the frequency domain, leveraging both frequency and time-domain features for anomaly detection.
    \item LLMAD~\cite{LLMAD}: An LLM-based method that prompts the LLM with time-series data, in-context learning examples, and contextual information for anomaly detection.
    \item SigLLM~\cite{SigLLM}: An LLM-based method in Detector mode and Prompter mode.
    The Detector mode prompts LLMs to predict the next steps in the time series and detects anomalies by comparing predictions with actual output, while the Prompter mode directly prompts the LLMs with the time-series data to identify anomaly indices.
\end{itemize}

\textbf{Datasets.}
We evaluate the proposed \sys{} on three datasets: KPI, Yahoo, and Internal.

\begin{itemize}
    \item KPI Dataset~\cite{kpidataset}: A real-world dataset with manually labeled incidents collected from five large internet companies, comprising 27 continuous metrics over several months.
    Due to the scarcity of anomalies in some metrics, we filter them out and use the remaining 19 metrics for evaluation.
    \item Yahoo Dataset~\cite{yahoodataset}: The dataset contains 367 real and synthetic time series across four partitions \texttt{A1 $\sim$ A4}, with anomalies exhibiting various characteristics, including spikes, trends, and seasonal changes.
    \item Internal Dataset: The dataset is collected from an AI training platform at \msft{} and includes 20 hardware metrics such as GPU utilization, GPU memory usage, CPU utilization, disk usage, and network traffic. 
\end{itemize}

We set the split ratio to 0.7 for each metric or partition in the datasets to ensure a similar anomaly ratio between the training and test sets.
For Yahoo dataset \texttt{A2} partition, we use a 0.5 split ratio to ensure both sets contain anomaly samples.
The chunk size is set to 2500 for the KPI dataset, 500 for the Yahoo dataset and 1000 for the Internal dataset.
We use one-for-one dataset mode for the KPI dataset, where \sys{} trains a set of anomaly detection rules for each metric.
We use one-for-all dataset mode for the Yahoo dataset and the Internal dataset, where \sys{} trains a set of rules for each partition and each hardware metric respectively.

\textbf{Test Machines.}
All LLM experiments are conducted on an Azure Standard\_D16as\_v4 VM~\cite{d16asv4} with 16 vCPUs and 64 GiB memory, while all model training and evaluation are performed on an Azure Standard\_ND96amsr\_A100\_v4 VM~\cite{nd96amsrv4} with 96 vCPUs and 8 NVIDIA A100 GPUs.

\textbf{LLM Endpoints.}
We use the Azure OpenAI service~\cite{aoai} to access the endpoint of LLM models~\cite{aoai-models}, including GPT-3.5~\cite{gpt3}, GPT-4-32k~\cite{gpt4}, and GPT-4o~\cite{gpt4o}, for all LLM experiments in both baselines and \sys{}.

\begin{figure}[t]
    \centering
    \includegraphics[clip, trim=0 0 0 50, width=\linewidth]{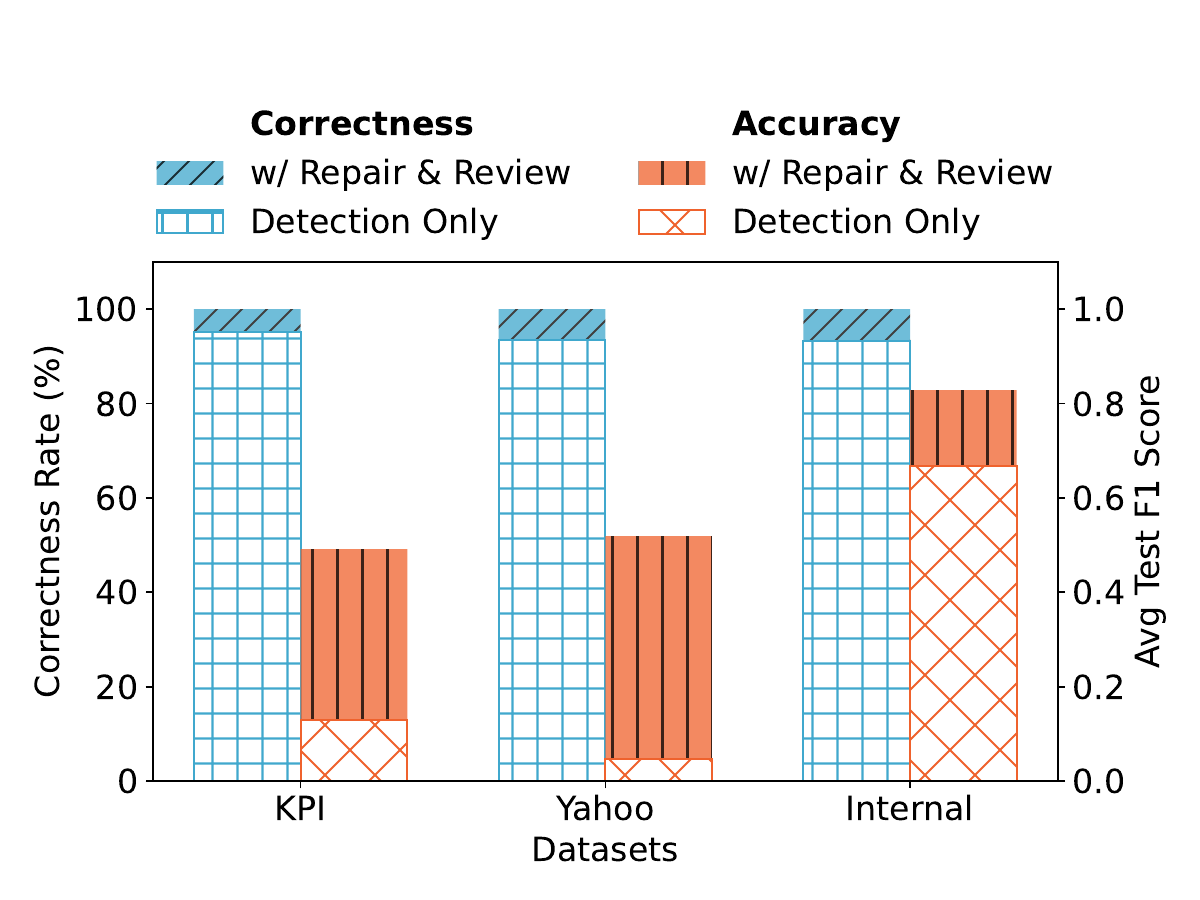}
    \caption{Comparison of the correctness rate and average test $F_1$ score of the Training Engine with only the Detection Agent versus full Training Engine with Repair and Review Agents.}
    \label{fig:evaluation-correctness-accuracy}
\end{figure}

\subsection{Correctness and Accuracy Improvement}
To demonstrate how feedback loops with the Repair and Review Agents improve the correctness and accuracy of the anomaly detection rules, we evaluate the Training Engine in two settings: with only the Detection Agent and with all three agents including the Detection, Repair, and Review Agents.
We run each metric or partition in the dataset for 50 trials with each trial containing 20 iterations.
In the Detection Agent only setting, at each iteration, the agent receives the time-series data along with the anomaly detection rule from the previous iteration.
For correctness rate, we measure how many anomaly detection rules generated in each trial have no syntax errors.
For accuracy, we calculate the average $F_1$ score on the test set for anomaly detection rules generated in all trials of all iterations.

\Cref{fig:evaluation-correctness-accuracy} shows the comparison of correctness rate and average test $F_1$ score between two settings.
We observe that the correctness rate of the Detection Agent only setting is 95.2\% in the KPI dataset, 93.5\% in the Yahoo dataset, and 93.3\% in the Internal dataset.
With the help of the Repair Agent, all anomaly detection rules generated by the Training Engine have no syntax errors.
In addition, the Review Agent helps improve the average test $F_1$ score of the anomaly detection rules by $3.8\times$ in the KPI dataset, $11.3\times$ in the Yahoo dataset, and $1.2\times$ in the Internal dataset, clearly demonstrating the effectiveness of feedback loops in the Training Engine.

\subsection{Accuracy Guarantee}

\begin{table}[t]
    \centering
    \caption{$F_1$ Score comparison on all metrics where \textbf{\sys{} w/o Aggregator} has regressions compared to baseline model.}
    \begin{tabular}{@{}lcccc@{}}
    \toprule
    \textbf{Metric} & \makecell{Baseline \\ Model} & \makecell{\textbf{\sys{} w/o} \\ \textbf{Aggregator}} & \textbf{\sys{}} \\ \midrule
    KPI-\texttt{02e99} & 0.99 & 0.96 \small{(-0.03)} & 0.99 \small{(+0.00)} \\
    KPI-\texttt{07927} & 0.99 & 0.67 \small{(-0.32)} & 0.99 \small{(+0.00)} \\
    KPI-\texttt{1c35d} & 0.89 & 0.80 \small{(-0.09)} & 0.99 \small{(+0.10)} \\
    \midrule
    Yahoo-\texttt{A2} & 0.90 & 0.87 \small{(-0.03)} & 0.95 \small{(+0.05)} \\
    Yahoo-\texttt{A4} & 0.85 & 0.84 \small{(-0.01)} & 0.87 \small{(+0.02)} \\
    \bottomrule
    \end{tabular}
    \label{tab:evaluation-accuracy-guarantee}
\end{table}

To demonstrate how model fusion provides an accuracy guarantee for the anomaly detection rules, we compare two versions of \sys{}, one without and one with the Aggregator module.
In the version without the Aggregator, we train the rules on all time-series data from the training set of each dataset, and during inference, the prediction labels are directly generated from the rules.
In the version with the Aggregator, we first select the deep learning model with the highest $F_1$ score on the training set as the base detector for each dataset, then collect false positives and false negatives in the training set from this base detector to train the anomaly detection rules.
\Cref{tab:evaluation-accuracy-guarantee} shows all metrics in three datasets where \sys{} without the Aggregator has accuracy regressions compared to the baseline.
In the KPI and Yahoo datasets, \sys{} without the Aggregator shows accuracy regressions on 3 and 2 metrics, respectively, with up to $32\%$ accuracy drop.
In contrast, \sys{} with the Aggregator has no accuracy regressions across any metrics in all three datasets.

\begin{figure*}[t]
    \centering
    \begin{subfigure}[b]{0.33\linewidth}
        \centering
        \includegraphics[width=\linewidth]{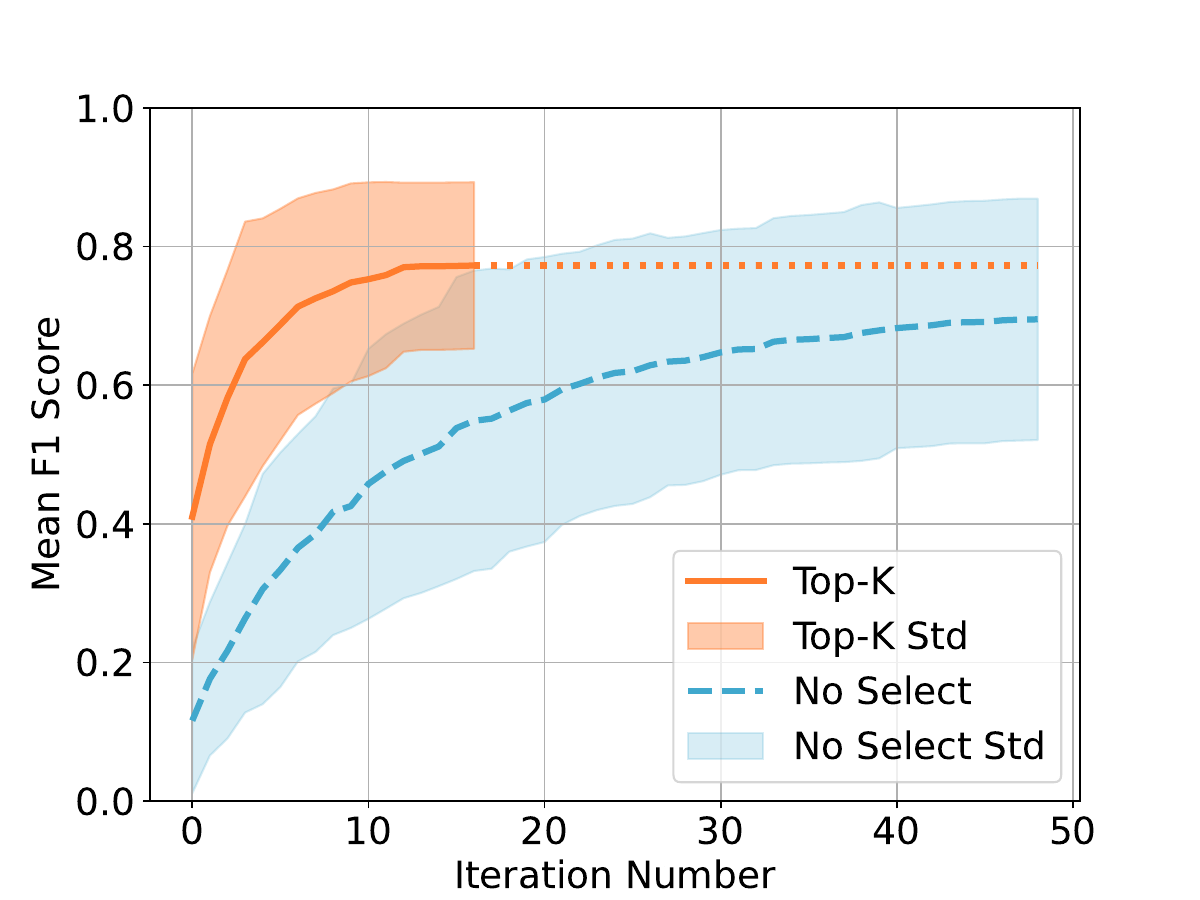}
        \caption{KPI Dataset}
        \label{fig:evaluation-top-k-kpi}
    \end{subfigure}
    \begin{subfigure}[b]{0.33\linewidth}
        \centering
        \includegraphics[width=\linewidth]{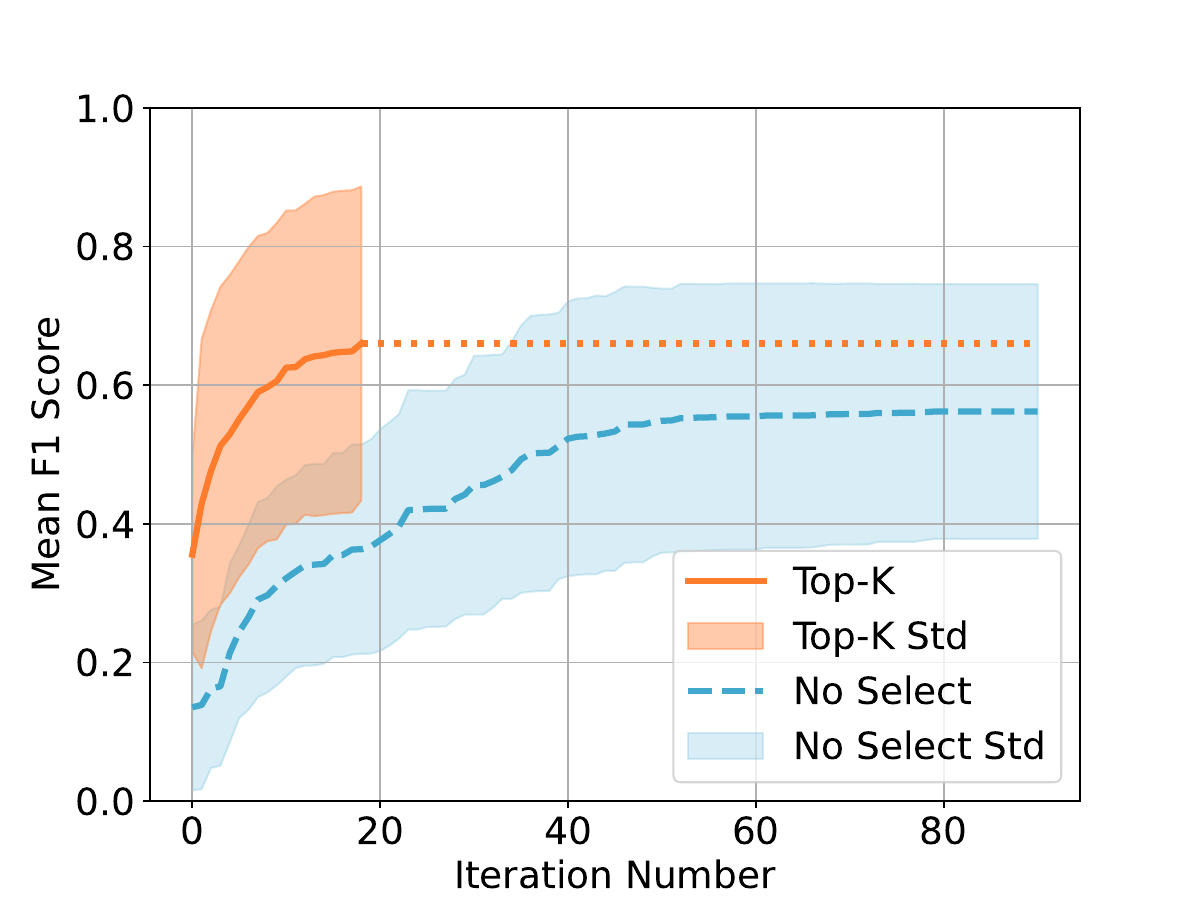}
        \caption{Yahoo Dataset}
        \label{fig:evaluation-top-k-yahoo}
    \end{subfigure}
    \begin{subfigure}[b]{0.33\linewidth}
        \centering
        \includegraphics[width=\linewidth]{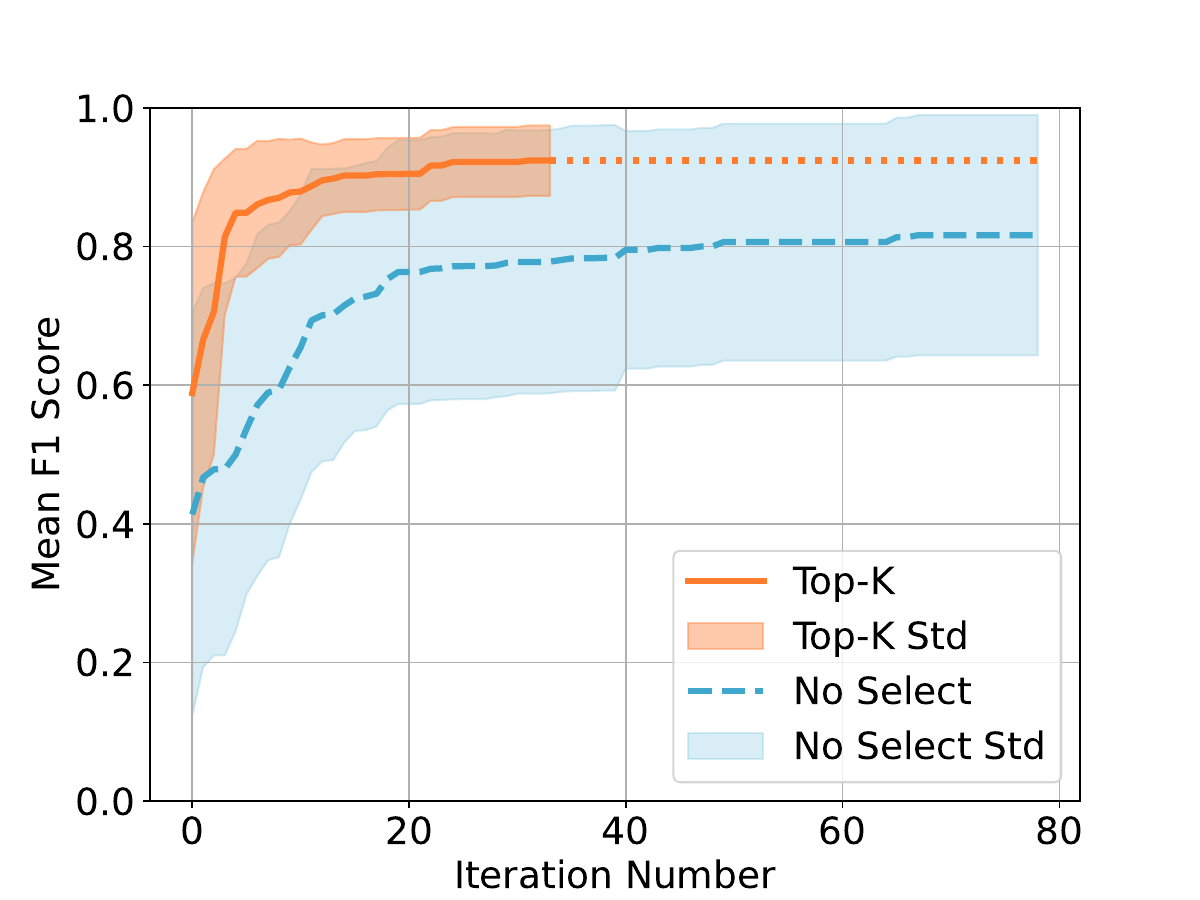}
        \caption{Internal Dataset}
        \label{evaluation-top-k-internal}
    \end{subfigure}
    \caption{Average $F_1$ score comparison on the training sets between top-$k$ rule selection and no selection.} %
    \label{fig:evaluation-top-k}
\end{figure*}

\begin{table*}[t]
    \centering
    \caption{Accuracy comparison of different anomaly detection methods on three datasets.}
    \begin{tabular}{l|c@{\hspace{1em}}c@{\hspace{1em}}c@{\hspace{1em}}c@{\hspace{1em}}c@{\hspace{1em}}c@{\hspace{1em}}c@{\hspace{1em}}c@{\hspace{1em}}c}
    \toprule
    \multirow{2}{*}{\textbf{Method}}      & \multicolumn{3}{c}{\textbf{KPI}} & \multicolumn{3}{c}{\textbf{Yahoo}} & \multicolumn{3}{c}{\textbf{Internal}} \\
           & \small{Precision} & \small{Recall} & $F_1$  & \small{Precision} & \small{Recall} & $F_1$ & \small{Precision} & \small{Recall} & $F_1$ \\
    \midrule
    AnomalyTransformer~\cite{AnomalyTransformer}    &   0.482 &	0.395 &	0.282 &	0.297 &	0.055 &	0.058 &	0.721 &	0.581 &	0.598 \\
    AutoRegression~\cite{AR}       & 0.731 &	0.699 &	0.668 &	0.508 &	0.623 &	0.526 &	0.735 &	0.618 &	0.582 \\
    FCVAE~\cite{FCVAE}       & 0.881 &	0.808 &	0.818 &	0.721 &	0.438 &	0.464 &	\textbf{0.944} &	0.479 &	0.581 \\
    LSTMAD~\cite{LSTMAD}       & \textbf{0.887} &	0.789 &	\textbf{0.819} &	0.425 &	0.493 &	0.350 &	0.865 &	\textbf{0.655} &	
    \textbf{0.724} \\
    TFAD~\cite{TFAD}       & 0.690 &	0.670 &	0.564 &	\textbf{0.830} &	\textbf{0.726} &	\textbf{0.773} &	0.783 &	0.500 &	0.588 \\
    \midrule
    LLMAD~\cite{LLMAD}       & 0.033 &	\textbf{0.940} &	0.053 &	0.090 &	0.687 &	0.142 &	0.623 &	0.575 &	0.537\\
    SigLLM - Detector~\cite{SigLLM}       & 0.001 &	0.239 &	0.002 &	0.004 &	0.543 &	0.009 &	0.668 &	0.486 &	0.443 \\
    SigLLM - Prompter~\cite{SigLLM}       & 0.007 &	0.498 &	0.014 &	0.006 &	0.027 &	0.010 &	0.178 &	0.424 &	0.188 \\
    \midrule
    \textbf{\sys{}}                 & 0.951 &	\textbf{0.859} &	\textbf{0.897} &	0.860 &	0.768 &	\textbf{0.810} &	0.955 &	\textbf{0.910} &	0.929 \\
    \textbf{~~~~-- w/ FN Rules Only} & 0.882 &	\textbf{0.859} &	0.860 &	0.793 &	\textbf{0.772} &	0.782 &	0.955 &	\textbf{0.910} &	0.929 \\
    \textbf{~~~~-- w/ FP Rules Only} & \textbf{0.997} &	0.785 &	0.864 &	\textbf{0.920} &	0.697 &	0.763 &	0.894 &	0.643 &	0.722 \\
    \textbf{~~~~-- w/o Aggregator}    & 0.954 &	0.850 &	0.890 &	0.901 &	0.719 &	0.800 &	\textbf{1.000} &	0.882 &	\textbf{0.936} \\
    \bottomrule
    \end{tabular}
    \label{tab:evaluation-main-results}
\end{table*}

\subsection{Efficiency Enhancement}

To measure the efficiency of the top-$k$ rule selection, we compare the accuracy of \sys{} and \sys{} with no selection.
In the no-selection setting, the Detection Agent proposes a single rule per iteration, which is then validated by the Repair and Review Agents.
In top-$k$ rule selection, the Detection Agent proposes 5 rules in parallel, all of which are validated, and the best-performing rule is selected.
In the next iteration, the Detection Agent proposes new 5 rules based on the best rule from the previous iteration.
We use GPT-4-32k as the LLM back-end for evaluation and run each metric or partition for 5 trials.
We compare the average $F_1$ scores on the training set for the generated rules across all trials, with both settings executed by \sys{} without the Aggregator.

\Cref{fig:evaluation-top-k} shows the average $F_1$ score across all trials and metrics for each dataset on the training set.
For a fair comparison, we evaluate the average $F_1$ score at points where the iteration number for no-selection is five times that of top-$k$ rule selection, as the latter generates 5 rules per iteration.
On the KPI dataset, the $F_1$ score for top-$k$ selection at iteration 8 is 0.736, outperforming 0.682 for no-selection at iteration 40.
On the Yahoo dataset, top-$k$ selection achieves an $F_1$ score of 0.597 at iteration 8, outperforming 0.523 for no-selection at iteration 40.
On the Internal dataset, the $F_1$ score for top-$k$ selection is 0.870 at iteration 8, outperforming 0.795 for no-selection at iteration 40.
Additionally, we observe that the converged $F_1$ scores at the end of the training are consistently higher with top-$k$ rule selection than with no-selection across all datasets.
This demonstrates that top-$k$ rule selection accelerates training and improves the accuracy of generated rules.

\subsection{End-to-End Results}

For end-to-end results, we compare \sys{} with all baselines across three datasets, calculating the average precision, recall, and $F_1$ score for each method on test sets.

For deep learning-based methods, we perform a grid search on hyperparameters for each dataset and select the configuration yielding the highest $F_1$ score on the training set.
For LLM-based methods, we evaluate 10 trials per metric on the test set and report the best accuracy across all trials.
To match the configurations in the original papers, LLMAD uses GPT-4-32k as the LLM back-end~\cite{LLMAD} while SigLLM uses GPT-3.5~\cite{SigLLM} in the baselines.

For \sys{}, we first select the baseline model with the highest $F_1$ score on the training set as the base detector.
We then train anomaly detection rules from the false negative samples (\sys{} w/ FN Rules Only) and false positive samples (\sys{} w/ FP Rules Only) of this base detector.
These two sets of rules are then fused with the base detector using~\Cref{alg:aggregation} to generate the final results.
We train each metric for 5 trials and report the best accuracy across all trials.
GPT-4o is used as the LLM back-end for \sys{}.

\Cref{tab:evaluation-main-results} shows the overall accuracy comparison of all methods across the three datasets.
\sys{} outperforms all baselines in every dataset.
In the KPI dataset, \sys{} achieves an $F_1$ score of 0.897, which is \maxspeedupkpi higher than the best baseline, LSTMAD.
In the Yahoo dataset, \sys{} achieves an $F_1$ score of 0.810, \maxspeedupyahoo higher than the best baseline, TFAD.
In the Internal dataset, \sys{} achieves an $F_1$ score of 0.936, \maxspeedupprivate higher than the best baseline, LSTMAD.
\sys{} w/ FN Rules Only achieves the highest recall in all datasets, demonstrating the effectiveness of addressing false negatives in the base model, as recall measures the false negative rate.
Similarly, \sys{} w/ FP Rules Only achieves the highest precision in the two public datasets.

\subsection{Cost Analysis}

\begin{table}[t]
    \centering
    \caption{Cost analysis of rule training per iteration.}
    \begin{tabular}{@{}lcccc@{}}
    \toprule
    \textbf{Category} & \textbf{KPI} & \textbf{Yahoo} & \textbf{Internal} \\ \midrule
    Input Tokens & 38238 & 39064 & 70754 \\
    Output Tokens & 1119 & 1669 & 909 \\
    \midrule
    Input API Budget (USD) & 0.096 & 0.098 & 0.177 \\
    Output API Budget (USD) & 0.011 & 0.017 & 0.009 \\
    \bottomrule
    \end{tabular}
    \label{tab:evaluation-api-cost}
\end{table}

\begin{table}[t]
    \centering
    \caption{Inference latency (seconds) comparison of different methods.}
    \begin{tabular}{@{}lcccc@{}}
    \toprule
    \textbf{Method}               & \textbf{KPI} & \textbf{Yahoo} & \textbf{Internal} \\  \midrule
    AnomalyTransformer~\cite{AnomalyTransformer}    &        171.23	& 171.58 & 	53.38               \\
    AutoRegression~\cite{AR}          &  0.68 &	0.88 &	0.26                \\
    FCVAE~\cite{FCVAE}      &     20.04 &	25.67 & 	7.82               \\
    LSTMAD~\cite{LSTMAD}        &   2.92 &	3.15 &	0.90     \\
    TFAD~\cite{TFAD}           & 49.19 & 	67.43 &	19.08      \\
    \midrule
    \textbf{\sys{} w/o Aggregator}            & 0.97                    & 1.97 & 0.59                 \\
    \bottomrule
    \end{tabular}
    \label{tab:evaluation-runtime-latency}
\end{table}

\textbf{API Budget for Training.}
\Cref{tab:evaluation-api-cost} shows the input and output token counts for each dataset.
We collect these statistics from the training of anomaly detection rules using false negative and false positive samples.
The average token count per iteration is calculated across all trials for each dataset.
Using GPT-4o as the LLM backend, the cost of input tokens per iteration is up to \$0.177, while the cost of output tokens per iteration is up to \$0.017.
For a total of 50 iterations, the cost of training the anomaly detection rules is under \$10 for all datasets in total.

\textbf{Runtime Latency for Inference.}
\Cref{tab:evaluation-runtime-latency} compares the inference latency of \sys{} with the baselines across three datasets.
We perform inference on the test set and calculate the average latency for all metrics or partitions in the dataset.
All baselines and \sys{} are run on a Standard\_D16as\_v4 VM~\cite{d16asv4} using the CPU.
\sys{} achieves low inference latency compared to the baselines, as the anomaly detection rules, implemented in Python, are lightweight.
Compared to the baseline with the highest average $F_1$ score on each dataset, \sys{} achieves a $3.0\times$ speedup on the KPI dataset, a $34.3\times$ speedup on the Yahoo dataset, and a $1.5\times$ speedup on the Internal dataset.

\subsection{Case Study}

\begin{figure}[t]
    \centering
    \begin{subfigure}[b]{.95\linewidth}
        \centering
        \includegraphics[width=\linewidth]{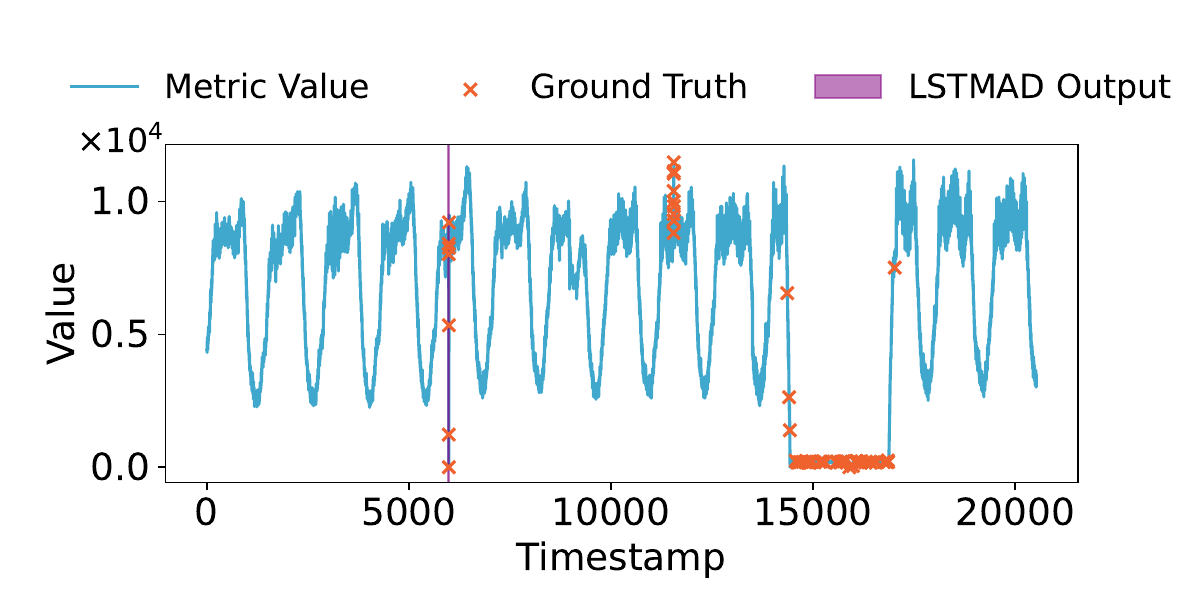}
        \caption{LSTMAD}
        \label{fig:evaluation-case-study-lstmad}
    \end{subfigure}
    \begin{subfigure}[b]{.95\linewidth}
        \centering
        \includegraphics[width=\linewidth]{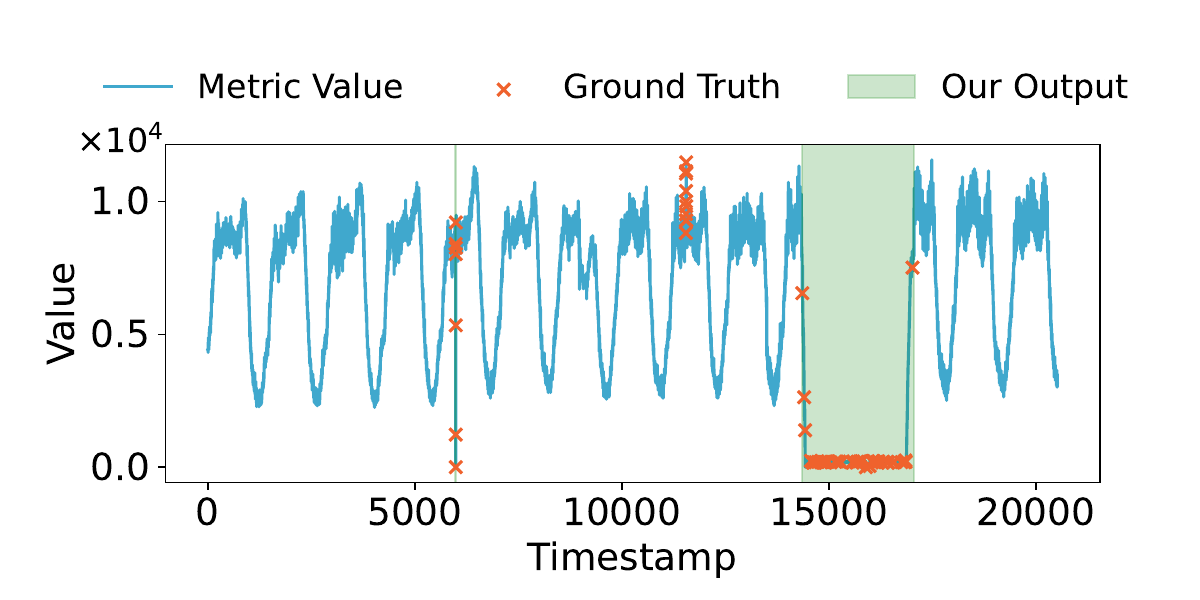}
        \caption{\sys{}}
        \label{fig:evaluation-case-study-adara}
    \end{subfigure}
    \caption{Comparison of \sys{}'s output with the LSTMAD baseline on the \texttt{e0770} metric in KPI dataset.}
    \label{fig:evaluation-case-study}
\end{figure}

\begin{figure}[t]
    \centering
    \begin{subfigure}[t]{\linewidth}
        \centering
        \begin{lstlisting}[language=Python,style=customnew]
labels = np.zeros(sample.shape[0], dtype=int)
# Abnormal Rule 1: If there are sudden spikes 
# or drops in values.
spikes_or_drops = np.abs(np.diff(sample[:, 0]))
labels[1:][spikes_or_drops > 4000] = 1
# Abnormal Rule 2: If there is a prolonged 
# period (e.g., more than 10 consecutive 
# points) of extreme values.
extreme_values = (sample[:, 0] < 2000) | 
    (sample[:, 0] > 15000)
for i in range(len(extreme_values) - 10):
    if np.all(extreme_values[i:i+10]):
        labels[i:i+10] = 1
\end{lstlisting}
    \end{subfigure}
    \vspace{-5pt}
    \caption{Anomaly detection rules trained from false negative samples of LSTMAD on metric \texttt{e0770} in KPI dataset.}
    \label{lst:evaluation-case-study-rule}
\end{figure}

In this case study, we demonstrate how the anomaly detection rules trained by \sys{}, based on an existing anomaly detection model, can improve model accuracy.
We choose the \texttt{e0770} metric from the KPI dataset as an example, where the best baseline model, LSTMAD, has a low recall of 0.33 on the test set.
\Cref{fig:evaluation-case-study-lstmad} shows LSTMAD's output on the test set.
There are 3 ground-truth anomalies: two spikes and one prolonged period of extreme values, while LSTMAD detects only the spike around timestamp 5,000, missing the other two anomalies.

We train anomaly detection rules using the false negative samples from LSTMAD on the training set in \sys{}.
\Cref{lst:evaluation-case-study-rule} shows the rules generated by \sys{}, which consist of two conditions: the first detects sudden spikes or drops, while the second identifies prolonged periods of extreme values.
\Cref{fig:evaluation-case-study-adara} shows \sys{}'s output on the test set, where the results are aggregated from LSTMAD and the rules using the Aggregator module.
Particularly, the second condition helps \sys{} detect a missed anomaly around timestamp 15,000, improving recall from 0.33 to 0.67.
It is also worth noting that a false negative still persists around timestamp 12,000 in \sys{}'s output.
This could be due to the minor deviation behavior in this sample not being present in the training set, or the threshold for detecting spikes in the rules may require further tuning.

\section{Related Works}
\textbf{Time-Series Anomaly Detection.}
Prior works on time-series anomaly detection are primarily deep learning-based~\cite{AnomalyTransformer,AR,Donut,FCVAE,Informer,LSTMAD,SPOT,SRCNN,TFAD,TranAD} and rule-based methods~\cite{ServiceLab,FBDetect,Resin}.
Early deep learning-based models focus on directly predicting the next value based on the observation from current window~\cite{AR,LSTMAD}.
With the introduction of encoder-decoder and variational encoder, later models reconstruct the time-series signal by denoising the input and flag anomalies if the observed signal deviates largely from the reconstructed signal~\cite{Donut,FCVAE,SRCNN,TFAD,TranAD,AnomalyTransformer}.
Compared to deep learning-based methods, rule-based methods are more prevalent in industry due to their explainability.
Resin~\cite{Resin} traces memory usage and reports an anomaly if the usage of the current period exceeds mean plus three times the standard deviation of the previous period.
Similarly, FBDetect~\cite{FBDetect} enables users to establish a detection threshold that defines the required magnitude of metric deviation for it to be classified as an anomaly.

\textbf{LLMs for Anomaly Detection.}
More recently, the emergence of LLMs have led to their applications in anomaly detection.
There have been works focusing on leveraging LLMs to perform log-based anomaly detection~\cite{logllm,liu2024logprompt,egersdoerfer2023early,le2021log}.
For example, NeuralLog~\cite{le2021log} transforms log messages into semantic vectors and uses a Transformer-based model to classify anomalies.
LogLLM~\cite{logllm} preprocesses logs with regular expressions, employs BERT for semantic extraction, and uses an LLM for sequence classification.
Some works have also studied using LLMs for time-series anomaly detection~\cite{LLMAD,SigLLM,LLM4TSAD?}.
They either fine-tune the LLMs based on time-series data or directly prompt the LLMs for output labels.
To the best of our knowledge, our work is the first to leverage LLMs to generate explainable and reproducible detection rules for time-series anomaly detection.

\textbf{LLMs for Cloud Reliability.}
Apart from cloud monitoring, LLMs have also been applied in other aspects of cloud reliability from root cause analysis to incident mitigation~\cite{RCACopilot,iclrca,cloudatlas,wang2024rcagent,goel2024xlifecycle,jiang2024xpert,ahmed2023recommending}.
RCACopilot~\cite{RCACopilot} is an on-call system for automated cloud incident root cause analysis, combining customizable incident handler workflows and LLM-based root cause prediction to streamline diagnostic data collection.
\cite{iclrca} proposes an in-context learning method for root cause analysis using LLMs without fine-tuning, employing historical incidents as examples to inform the model, outperforming fine-tuned models in accuracy and utility.
Atlas~\cite{cloudatlas} is a tool leveraging large language models to automate the transformation of unstructured system information into structured causal graphs, enhancing fault localization in cloud systems by constructing high-quality causal representations.

\vfill
\section{Conclusion}

Production anomaly detection systems require three indispensable properties at the same time: explainability, reproducibility, and autonomy.
In this paper, we present \sys{}, a time-series anomaly detection system that autonomously generates rules via LLMs to detect anomalies.
By leveraging LLMs' capabilities in time-series understanding and code generation, \sys{} ensures that the rules are both explainable and reproducible.
\sys{} comprises an agent-based pipeline that iteratively proposes, repairs, and reviews anomaly detection rules to ensure their quality.
Additionally, \sys{} incorporates model fusion for accuracy guarantees and performs top-$k$ rule selection to enhance efficiency.
Our evaluation on both public and internal datasets demonstrates that \sys{} outperforms state-of-the-art time-series anomaly detection methods, achieving up to a \maxspeedupprivate improvement in $F_1$ score.

\break

\balance
\bibliographystyle{plain}
\bibliography{paper}

\clearpage
\nobalance
\onecolumn
\section*{Appendices}
\appendix

\section{Prompts}
\label{sec:appendix-prompts}

\subsection{Detection Agent}
\Cref{fig:design-detection-agent-prompt} shows the prompt template used by the Detection Agent in \sys{}.
The template begins with a high-level task summary for the Detection Agent, followed by step-by-step instructions on the data format and function invariants.
Finally, we provide important notes that we empirically found useful for the Detection Agent to follow to improve the accuracy of anomaly detection rules.

\begin{figure*}[h]
\begin{tcolorbox}[colback=transparentblue, colframe=lightblue, boxrule=0.5pt, arc=2pt, width=\linewidth, title=Prompt Template for the Detection Agent, fonttitle=\bfseries\color{black}, colbacktitle=lightblue, coltitle=black]

\textbf{Task Summary:} \\
You are an AI assistant that helps people write detection rules to determine whether a piece of time series data is abnormal (negative) or not (positive). The time series data is collected during a task for cloud service $\ldots$

\hfill\break
\textbf{Step-by-step Instructions:} \\
You should achieve the task in the following steps:
\begin{enumerate}
    \item You will be given the data sample in the following format: $\ldots$
    \item You should give a Python function \texttt{inference(sample: np.ndarray) -> labels: np.ndarray} to write various rules to describe the pattern of given negative/abnormal samples and exclude all given positive/normal samples.
    \item The function will take a sample of numpy array with shape $(X, 2)$ as input, where each row is a tuple of (value, index). You should return the labels as an \texttt{np.ndarray} of shape $(X,)$, and for each index, \texttt{value=1} means the data of the index is abnormal, and \texttt{value=0} means the data of the index is normal.
    \item Beyond anomalies, you can describe how normal data behave in comments, in the format of \texttt{``Normal Rule 1 \textbackslash n Normal Rule 2 ...''} Ideally, if the inference function returns no abnormal indices, then the data MUST satisfy all normal rules you describe in comments. You should strictly use the following format:
    $\ldots$
\end{enumerate}

\hfill\break
\textbf{Important Notes:}
\begin{enumerate}
    \item Your code should not hard code any information about the label given to you in example data.
    \item You should not hard code the indices of anomalies.
    \item You should make sure the python code is correct and can be executed without any error.
    \item If your code uses any external libraries, you should include the import statements in the code.
\end{enumerate}

\end{tcolorbox}
\caption{A prompt template used by the Detection Agent in \sys{}.}
\label{fig:design-detection-agent-prompt}
\end{figure*}

\newpage
\subsection{Repair Agent}
\Cref{fig:design-repair-agent-prompt} shows the prompt template used by the Repair Agent in \sys{}.
If the code contains syntax errors, the Repair Agent is provided with error messages and the incorrect rule.
The Repair Agent suggests a revised version of the rules until all syntax errors are fixed.

\begin{figure*}[h]
\begin{tcolorbox}[colback=transparentblue, colframe=lightblue, boxrule=0.5pt, arc=2pt, width=\linewidth, title=Prompt Template for the Repair Agent, fonttitle=\bfseries\color{black}, colbacktitle=lightblue, coltitle=black]

\textbf{Task Summary:} \\
You are an AI assistant that fixes syntax and runtime errors in Python code. You will be given a Python code snippet along with the error message indicating details for syntax and/or runtime errors. You should fix the errors in the code and make sure the code can be executed without any error $\ldots$

\hfill\break
\textbf{Step-by-step Instructions:} \\
You should achieve the task in the following steps:
\begin{enumerate}
    \item You will be given the data sample in the following format: $\ldots$
    \item You are given a python function \texttt{inference(sample: np.ndarray) -> labels: np.ndarray} to write various rules to describe and remember the pattern of given negative/abnormal samples and exclude all given positive/normal samples. The function will take a sample of numpy array with shape $(X, 2)$ as input $\ldots$
\end{enumerate}

\hfill\break
\textbf{Important Notes:}
\begin{enumerate}
    \item You should output the fixed code following the same format as the input code, wrapping the code with \texttt{*** python begin ***} as the first line and \texttt{*** python end ***} as the last line. You must only use \texttt{*** python begin ***} and \texttt{*** python end ***} to wrap your fixed code for only once, don't use them for any other purpose.
    \item You should only focus on fixing the errors in the code and make sure the code can be executed without any error. You must not change other logic of the code unrelated to the errors.
\end{enumerate}

\end{tcolorbox}
\caption{A prompt template used by the Repair Agent in \sys{}.}
\label{fig:design-repair-agent-prompt}
\end{figure*}

\newpage
\subsection{Review Agent}
\Cref{fig:design-review-agent-prompt} shows the prompt template used by the Review Agent in \sys{}.
If the code has accuracy regression compared to the previous iteration or compared to the existing anomaly detection model, the Review Agent is provided with the current code, code difference with the last iteration if exists and performance metrics comparison. 
Additionally, the Review Agent is also provided with incorrect examples where the current code incorrectly predicts while the previous code or the model correctly predicts.
The Review Agent proposes a new set of rules until the performance regression is mitigated.

\begin{figure*}[h]
\begin{tcolorbox}[colback=transparentblue, colframe=lightblue, boxrule=0.5pt, arc=2pt, width=\linewidth, title=Prompt Template for the Review Agent, fonttitle=\bfseries\color{black}, colbacktitle=lightblue, coltitle=black]

\textbf{Task Summary:} \\
You are an AI assistant that reviews Python code changes and propose modifications. 
You will be given a Python code that contains various anomaly detection rules to describe and remember the pattern of given negative/abnormal samples and exclude all given positive/normal samples. 
The rules in this Python code will be used in combination with an anomaly detection model that performs anomaly detection. 
Both the rules and the anomaly detection model will generate anomaly labels, and the performance is a combination of anomaly labels from both sides, comparing against the ground-truth labels $\ldots$

\hfill\break
\textbf{Step-by-step Instructions:} \\
You should achieve the task in the following steps:
\begin{enumerate}
    \item You will be given the data sample in the following format: $\ldots$
    \item You are given a python function \texttt{inference(sample: np.ndarray) -> labels: np.ndarray} to write various rules to describe and remember the pattern of given negative/abnormal samples and exclude all given positive/normal samples $\ldots$
    \item The review process depends on the stage of rule generation.
    \item If previous code does not exist, that means we are at the first iteration for the rules. You will be given the current code for the rules. You will also be given the performance metrics of the current code combined with the anomaly detection model and the baseline performance from running only the anomaly detection model $\ldots$

    \item If previous code exists, that means we are iterating over the rules. You will be additionally given a code difference comparing the current code with the previous code. Our goal is make sure that the performance metric of the current code is better than the previous code.
    
    \item You will be given the performance metrics of the current code and either the baseline performance or performance from the previous code in the following format $\ldots$

    \item You will be given the code difference in the following format, if previous code exists $\ldots$
    
    \item You will also be given incorrect examples that the current code incorrectly predicts while the previous code or the anomaly detection model correctly predicts $\ldots$
    
\end{enumerate}

\hfill\break
\textbf{Important Notes:}
\begin{enumerate}
    \item You should output the fixed code following the same format as the input code, wrapping the code with \texttt{*** python begin ***} as the first line and \texttt{*** python end ***} as the last line. You must only use \texttt{*** python begin ***} and \texttt{*** python end ***} to wrap your fixed code for only once, don't use them for any other purpose.
    \item You should make sure the python code is correct and can be executed without any error.
    \item If your code uses any external libraries, you should include the import statements in the code.
\end{enumerate}

\end{tcolorbox}
\caption{A prompt template used by the Review Agent in \sys{}.}
\label{fig:design-review-agent-prompt}
\end{figure*}

\end{document}